\documentclass[sigconf]{acmart}

\usepackage{graphicx}
\usepackage{booktabs}
\usepackage{subcaption}
\usepackage{multirow}
\usepackage{bm}
\usepackage{colortbl}
\usepackage{listings}
\usepackage{float} 
\usepackage{nicematrix}  

\usepackage{makecell}
\usepackage{xcolor}
\usepackage{tikz}

\usepackage{geometry}
\usepackage{tcolorbox}
\tcbuselibrary{skins, breakable} 
\usepackage{enumitem} 
\usepackage{multicol}

\usepackage{CJKutf8}

\newtcolorbox{PromptBox}[1]{
	enhanced,                  
	title={#1},                
	colframe=blue!50!black,    
	colback=white,             
	colbacktitle=blue!10!white,
	coltitle=black,            
	fonttitle=\bfseries\sffamily\large, 
	boxrule=0.8pt,             
	arc=2mm,                   
	left=4mm, right=4mm, top=2mm, bottom=2mm, 
	drop fuzzy shadow,         
}

\definecolor{sneg3}{HTML}{E74C3C} 
\definecolor{sneg2}{HTML}{E67E22} 
\definecolor{sneg1}{HTML}{F1C40F} 
\definecolor{sneu0}{HTML}{1ABC9C} 
\definecolor{spos1}{HTML}{3498DB} 
\definecolor{spos2}{HTML}{2980B9} 
\definecolor{spos3}{HTML}{9B59B6} 

\newcommand{\sentimentdot}[1]{\tikz\draw[#1,fill=#1] (0,0) circle (0.5ex);}

\definecolor{goodgreen}{RGB}{46, 139, 87}  
\definecolor{badred}{RGB}{178, 34, 34}     
\newcommand{\cpos}[1]{\textcolor{goodgreen}{\scriptsize (+#1)}} 
\newcommand{\cneg}[1]{\textcolor{goodgreen}{\scriptsize (#1)}} 

\newcommand{\cbadpos}[1]{\textcolor{badred}{\scriptsize (#1)}}
\newcommand{\cbadneg}[1]{\textcolor{badred}{\scriptsize (+#1)}}

\newcommand{\best}[1]{\textbf{#1}}

\definecolor{graybg}{gray}{0.95}
\newcommand{\na}{\textcolor{gray}{--}}

\AtBeginDocument{%
  }

\copyrightyear{2026}
\acmYear{2026}
\setcopyright{cc}
\setcctype{by}
\acmConference[MM '26] {Proceedings of the 34th ACM International Conference on Multimedia}{November 10--14, 2026}{Rio de Janeiro, Brazil}
\acmBooktitle{Proceedings of the 34th ACM International Conference on Multimedia (MM '26), November 10--14, 2026, Rio de Janeiro, Brazil}
\acmISBN{979-8-4007-2213-4/2026/11}
\acmDOI{10.1145/3767308.3835866}

\begin{document}

\title[SentiLLM]{Semantic-Aligned Structural Abstraction for Multimodal Sentiment Analysis}

\author{Wei Chen}
\orcid{0009-0000-7820-3498}
\affiliation{%
	\institution{Huazhong Agricultural University}
	\city{Wuhan}
	\country{China}}
\email{weichen5498@webmail.hzau.edu.cn}

\author{Junkai Li}
\orcid{0009-0001-3236-6838}
\affiliation{%
	\institution{Huazhong Agricultural University}
	\city{Wuhan}
	\country{China}}
\email{junkaili@webmail.hzau.edu.cn}

\author{Tongguan Wang}
\orcid{0009-0001-8747-1790}
\affiliation{%
	\institution{Huazhong Agricultural University}
	\city{Wuhan}
	\country{China}}
\email{wang\_tg@webmail.hzau.edu.cn}

\author{Hui Liu}
\orcid{0009-0003-9920-1186}
\affiliation{%
	\institution{Huazhong Agricultural University}
	\city{Wuhan}
	\country{China}}
\email{liuhui\_1003@webmail.hzau.edu.cn}

\author{Feiyue Xue}
\orcid{0009-0007-0949-2987}
\affiliation{%
	\institution{Huazhong Agricultural University}
	\city{Wuhan}
	\country{China}}
\email{xuefeiyue@webmail.hzau.edu.cn}

\author{Chuanxiang Ma}
\orcid{0009-0005-7770-6911}
\affiliation{%
	\institution{Hubei University} 
	\city{Wuhan}
	\country{China}}
\email{mcx838@hubu.edu.cn}

\author{Ying Sha}
\orcid{0000-0002-6638-5009}
\authornote{Corresponding author.}
\authornote{Also with Engineering Research Center of Intelligent Technology for Agriculture, Hubei Engineering Technology Research Center of Agricultural Big Data, Key Laboratory of Smart Farming for Agricultural Animals.}
\affiliation{%
	\institution{Huazhong Agricultural University}
	\city{Wuhan}
	\country{China}} 
\email{shaying@mail.hzau.edu.cn}  

\renewcommand{\shortauthors}{Wei Chen et al.}

\begin{abstract}

	Multimodal Sentiment Analysis (MSA) aims to interpret complex human emotions by integrating natural language with non-verbal modalities. Non-verbal modalities share a structural isomorphism with natural language, as both can be viewed as feature sequences evolving over time. This isomorphism enables the transformation of non-verbal modalities into text-like tokens for unified semantic reasoning. Large Language Models (LLMs), designed to understand and generate sequential data, can thus be utilized to interpret complex affective sequences. However, existing LLM-based methods primarily capture low-level superficial features, failing to model affective semantics arising from structural variations and contextual interactions. To address this limitation, we propose \textbf{SentiLLM}, a unified framework that leverages \textit{Semantic-Aligned Structural Abstraction} to distill continuous raw signals into compact, semantically meaningful tokens. Specifically, we introduce a \textit{Dual-Stream Salience-Context Calibration Mechanism}, which  disentangles non-verbal feature sequences into a focus stream and an ambient stream. The focus stream captures salient sentiment shifts (e.g., facial expressions) guided by textual priors, while the ambient stream characterizes stable background states. Through calibrating these dynamic sentiment shifts against background states, SentiLLM effectively projects non-verbal modalities into a unified semantic space, making them naturally understandable for LLMs. Serving as a plug-and-play module, SentiLLM significantly improves discriminative performance with only a small number of trainable parameters. Our method achieves superior performance on four datasets—MOSI, MOSEI, CH-SIMS, and CH-SIMS v2—demonstrating the effectiveness of the structural abstraction paradigm in MSA. Our code is available at: \href{https://github.com/especiallyW/SentiLLM}{https://github.com/especiallyW/SentiLLM}.

\end{abstract}




\maketitle

\section{Introduction}\label{sec1}

	Multimodal Sentiment Analysis (MSA) \cite{Zhang2024ACS} leverages heterogeneous data sources—video, audio, and text—to achieve precise reasoning about human sentiment states. Driven by its pivotal value in business decisions and human-computer interaction, the field has garnered significant attention. Existing methods \cite{Hazarika2020MISAMA,li2023decoupled,tsai2019multimodal,wang2025dlf,luo2025towards,wang2025rclmufn} typically adopt a modality decoupling and fusion paradigm. These methods focus on designing complex network architectures to capture shared and specific modal information. However, over-reliance on specialized architectures constrains scalability and generalization, hindering the effective utilization of large-scale pre-trained knowledge.
	
	The advent of Multimodal Large Language Models (MLLMs) \cite{li2023blip,liu2024improved} necessitates a re-examination of the intrinsic nature of different modalities. Non-verbal modalities, such as video and audio, share a structural isomorphism with natural language, both can be viewed as feature sequences evolving over time \cite{sun2019videobert,hsu2021hubert,tsai2019multimodal}. For instance, video frames resemble words in a sentence, where semantics are defined by their predecessors and successors. Consequently, Large Language Models (LLMs) \cite{dubey2024llama,Yang2024Qwen25TR}, serving as sequence modeling experts embedded with vast world knowledge, possess an innate potential to process non-verbal sequences and perform unified reasoning.
	
	
	However, the direct application of LLMs to MSA faces a substantial \textit{semantic gap}. Raw audio-visual signals appear as continuous and redundant long sequences. A simple linear mapping often capture only surface-level features (e.g., facial textures or acoustic frequencies), failing to extract subtle sentiment cues, such as an angry tremble or ironic tone. This appearance-semantic misalignment prevents transformed features from aligning within the LLM semantic space. Therefore, a semantic translation mechanism is required to convert non-verbal modalities into LLM-comprehensible sentiment tokens.
	
	We argue that sentiment is not a static time step mapping. Instead, it represents the structural unity of salient semantic shifts and contextual states across time. We proposed that sentiment judgment follows \textit{\textbf{a Focus-Ambient Calibration Mechanism:}} \textit{\textbf{human first attend to salient sentiment shifts (the Focus), such as a sudden scream or raised eyebrow, and subsequently verify reliability through contextual cues (the Ambient)}}. Based on this, an ideal sentiment token must meet two criteria: (1) \textbf{Structural}, representing a significant dynamic process rather than a static snapshot; and (2) \textbf{Calibration}, verified by context to ensure semantic stability rather than random noise.
	
	
	Guided by these cognitive insights, we introduce SentiLLM, a parameter-efficient framework built on \textbf{Semantic-Aligned Structural Abstraction}—a process that distills continuous, unaligned raw signals into compact, semantically meaningful tokens. Transcending traditional feature extraction paradigms, SentiLLM functions as a spatial-temporal semantic translator, converting non-verbal signals into sentiment tokens. Specifically, to instantiate this theoretical calibration, we design a Dual-Stream Salience-Context Calibration Mechanism. This mechanism disentangles audio-visual signals into a \textbf{Focus Stream} carrying sentiment shifts and an \textbf{Ambient Stream} depicting contextual states. The focus stream utilizes textual semantic anchors to retrieve salient dynamic sentiment shifts, while the ambient stream captures global background states to calibrate focus information stability. Through this interaction, raw signals are abstracted into compact, semantically rich sentiment tokens. Benefiting from this structural abstraction, the generated sentiment tokens align seamlessly with the textual semantic space, making them compatible with LLM reasoning without requiring complex architectural modifications.
	

	To validate our method, we conducted extensive experiments on four benchmark datasets: MOSI, MOSEI, CH-SIMS, and CH-SIMS v2. Experimental results demonstrate that our method achieves competitive performance without complex architecture design, requiring only a small number of trainable parameters. Our main contributions are summarized as follows:
	
	\begin{itemize}
		\item[$\bullet$] Leveraging the structural isomorphism between non-verbal modalities and natural language, we propose SentiLLM, demonstrating the feasibility of unifying multimodal sentiment reasoning within the LLM semantic space.		
		\item[$\bullet$] Rooted in the \textit{Focus-Ambient Calibration}, we introduce a Dual-Stream Salience-Context Calibration Mechanism. This mechanism achieves the structural abstraction from raw signals to compact sentiment tokens suitable for LLMs.
		\item[$\bullet$] Extensive experiments on four datasets demonstrate the effectiveness of our proposed method. We also release our code to facilitate further community research.
	\end{itemize}

\section{Related Work}\label{sec2}

	\subsection{Multimodal Sentiment Analysis}\label{sec21}

		Traditional approaches prioritize intricate fusion mechanisms to capture inter-modal interactions. Early works \cite{Zadeh2017TensorFN,liu2018efficient,zadeh2018memory} explicitly model interactions via mathematical outer products. With the rise of Transformers \cite{vaswani2017attention} and Graph Neural Networks \cite{scarselli2008graph}, some methods \cite{zadeh2018multimodal,williams2018recognizing,tsai2019multimodal} introduce dynamic fusion graphs and architectural innovations to handle long-range dependency issues in unaligned sequences. Similarly, \cite{yu2021learning,han2021improving,rahman2020integrating} optimize fusion characteristics through learning paradigms and theory. Specifically, Self-MM \cite{yu2021learning} employs self-supervised learning for multi-task training, while MMIM \cite{han2021improving} maximizes mutual information to retain key data. MAG-BERT \cite{rahman2020integrating} integrates BERT \cite{devlin2019bert} with multimodal inputs. 
		
		Recently, text-dominant tasks have highlighted the importance of modality disentanglement \cite{liu2022make,Hazarika2020MISAMA,li2023decoupled,sun2022cubemlp,sun2024mfm,zhang2025modal}. These methods differentiate between specific and shared modalities to enhance robustness. For instance, DMD \cite{li2023decoupled} refines representation purity via distillation. DLF \cite{wang2025dlf} separates shared and specific information for hierarchical prediction. KAN-MCP \cite{luo2025towards} addresses modality imbalance caused by density differences using KANs \cite{liu2024kan}.
		
		However, these methods rely on complex, task-specific architectures. In this paper, we treat video and audio modalities as "temporal sentences", leveraging the extensive sequential knowledge embedded in pre-trained LLMs to capture robust sentiment shifts.

	\subsection{Large Language Models}\label{sec22}

		Recent advancements in LLMs have revolutionized sequence modeling and reasoning, such as GPT \cite{brown2020language}, LLaMA \cite{dubey2024llama}, and Qwen \cite{Yang2024Qwen25TR}. Through visual instruction tuning, some studies extended LLMs to vision-language tasks. Models like LLaVA \cite{liu2024improved}, Flamingo \cite{alayrac2022flamingo}, and BLIP-2 \cite{li2023blip} employ MLP or Q-Former as connectors, effectively bridging the gap between vision encoders and LLMs.
		
		In the MSA domain, some methods \cite{hu2022unimse,li2023unisa} reframe sentiment analysis as sequence generation, converting multimodal inputs into discrete tokens for unified generation. MSE-Adapter \cite{yang2025mse} designs a lightweight adapter, using a text-guide-mixer to filter non-text features for efficient adaptation.
		
		However, these methods typically resort to coarse linear mapping or instruction tuning, thereby failing to capture the intrinsic essence of sentiment. They treat audio-visual signals as static context supplements, failing to explicitly capture structural variations across time (e.g., sudden sentiment shifts). This absence of a denoising-calibration mechanism renders the model susceptible to environmental noise, limiting fine-grained reasoning capabilities. In this paper, we propose a Dual-Stream Salience-Context Calibration Mechanism. It actively extracts and calibrates salient sentiment shifts, achieving feature translation into the LLM semantic space. 

		\begin{figure*}[tp]
			\centering
			\includegraphics[width=1.0\textwidth]{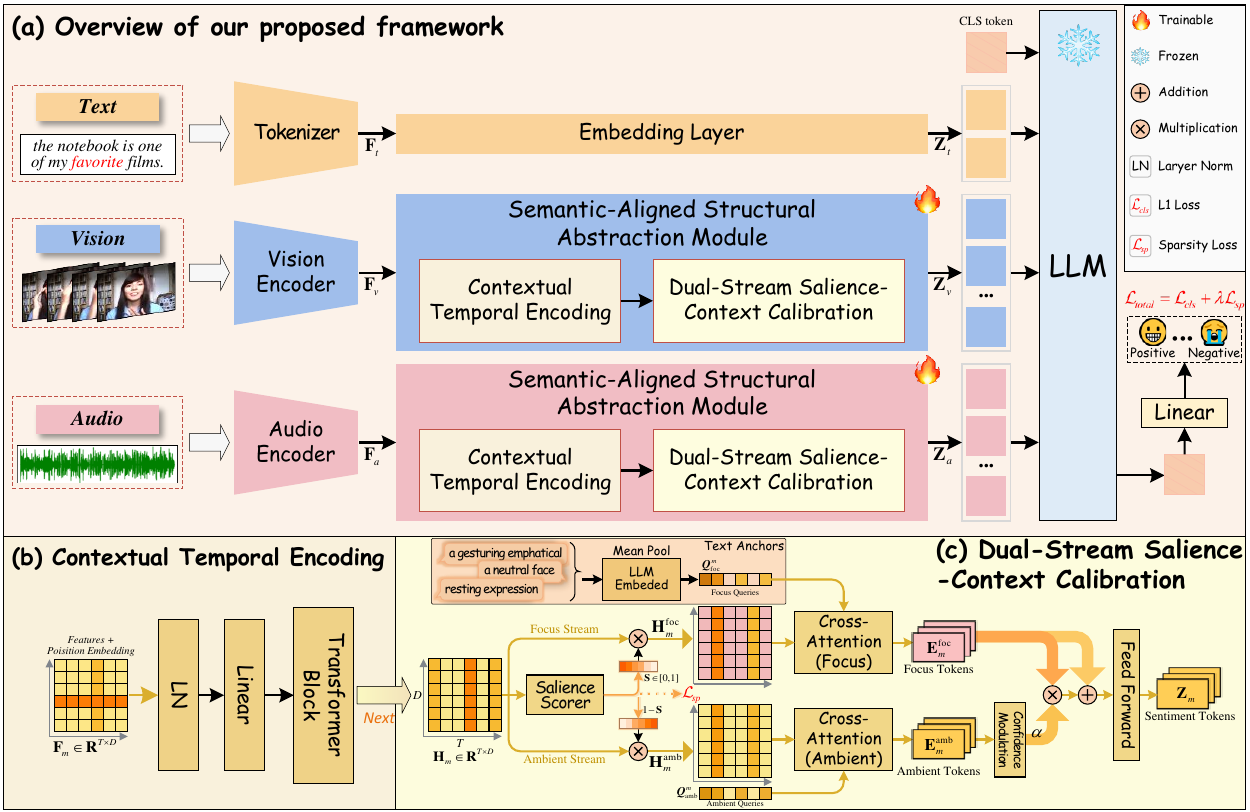}
			\caption{Overall architecture of SentiLLM. The model consists of two core modules: an Contextual Temporal Encoding Module, and a Dual-Stream Salience-Context Calibration Mechanism.}
			\label{fig:med:model}
		\end{figure*}

	\section{Methodology}\label{sec3}
		
		In this section, we detail the SentiLLM framework. First, we outline the overall architecture in Section~\ref{sec31}. Then, we describe the contextual temporal encoding module in Section~\ref{sec32}. Next, we elaborate on the dual-stream salience-context calibration mechanism in Section~\ref{sec33}. Finally, we define the joint optimization objectives in Section~\ref{sec34}.
		
		\subsection{Overall Architecture}\label{sec31}
		
			\noindent\textbf{Problem definition.} Given a multimodal dataset $\mathcal{D}$ with $N$ samples, each sample consists of text $\mathbf{X}_t$, visual $\mathbf{X}_v$, and audio $\mathbf{X}_a$ inputs alongside a sentiment label $y$. 
			
			Figure~\ref{fig:med:model} (a) illustrates the SentiLLM architecture. Specifically, for text input $\mathbf{X}_t$, we utilize a pre-trained tokenizer to obtain $\mathbf{F}_t$ and an embedding layer to transform it into vectors $\mathbf{Z}_t \in \mathbb{R}^{T \times D}$. For visual $\mathbf{X}_v$ and audio $\mathbf{X}_a$ inputs, we employ specific encoders to extract raw feature sequences $\mathbf{F}_v$ and $\mathbf{F}_a$, respectively. Subsequently, \textbf{the Semantic-Aligned Structural Abstraction Module} processes these non-verbal modalities. This module involves two stages: (1) \textbf{a contextual temporal encoding module} that transforms raw features to context-aware sequences; and (2) \textbf{a dual-stream salience-context calibration mechanism} that abstracts continuous sequences into compact, calibrated sentiment tokens $\mathbf{Z}_v$ and $\mathbf{Z}_a$ via soft salience disentanglement and dual-query abstraction. Finally, we concatenate these tokens with text features to form a unified sequence for LLM reasoning.		
		
		\subsection{Contextual Temporal Encoding}\label{sec32}
		
			The raw features $\mathbf{F}_m$ ($m \in \{v,a\}$) contain rich low-level information, but they lack temporal causality and contextual dependencies. To empower the model with temporal context awareness, we perform temporal modeling and positional enhancement.
			
			To achieve this, instead of using simple pooling operations, we employ a standard transformer block ($\text{TF}_{\text{Block}}$) to capture long-range temporal dependencies. Additionally, since subsequent disentanglement may disrupt local structure, we introduce absolute position embeddings ($\text{PE}$) to preserve sequential priors. Formally: 
			
			\begin{equation}
				\mathbf{H}_m = \text{TF}_{\text{Block}}(\text{Linear}(\text{LN}(\mathbf{F}_m)) + \text{PE})
			\end{equation}
			
			Here, $\text{Linear}(\cdot)$ denotes a linear mapping layer and $\text{LN}(\cdot)$ represents Layer Normalization \cite{ba2016layer}. This transformation yields context-aware features $\mathbf{H}_m \in \mathbb{R}^{T \times D}$, establishing a semantic foundation for the dual-stream mechanism.

		\subsection{Dual-Stream Salience-Context Calibration}\label{sec33}
		
			We face a core challenge after obtaining context-aware features: sequence redundancy. Some segments carry salient sentiment shifts (e.g., fleeting expressions), while others provide necessary background states. To filter redundancy, we propose the Dual-Stream Salience-Context Calibration mechanism to model both salient semantic shifts and contextual states. 
		
			\subsubsection{Soft Salience Disentanglement}\label{sec331}
		
				To distinguish sentiment shifts from background states, we require time-step level disentanglement. We design a soft salience disentanglement mechanism, instead of using hard truncation strategies. It learns step-wise salience probabilities, dynamically assigning features $\mathbf{H}_m$ to either the focus stream carrying salient actions or the ambient stream carrying background states.
				
				Specifically, we use a lightweight salience scoring network to calculate the probability score $S_t$ that the $t$-th time step represents a significant moments:
				
				\begin{equation}
					\mathbf{S} = \sigma (\mathbf{W}_2 \cdot \text{GELU}(\mathbf{W}_1 \mathbf{H}_m + \mathbf{b}_1) + \mathbf{b}_2)
				\end{equation}
				Here, $\mathbf{S} \in \mathbb{R}^{T \times 1}$ denotes the salience scores, $S_t \in [0,1]$, and $\sigma$ is the sigmoid function. Then we use $\mathbf{S}$ as a soft mask for weighted disentanglement:
				\begin{equation}
					\begin{aligned}
						\mathbf{H}_m^{\text{foc}} &= \mathbf{H}_m \odot \mathbf{S} \\ 
						\mathbf{H}_m^{\text{amb}} &= \mathbf{H}_m \odot (1 - \mathbf{S}) \\ 
					\end{aligned}
				\end{equation}
				where $\odot$ represents the Hadamard product. In this way, $\mathbf{H}_m^{\text{foc}}$ focuses on high-salience dynamic sentiment shifts, while $\mathbf{H}_m^{\text{amb}}$ retains low-salience contextual background information. This preserves the complete temporal length and positional information, achieving adaptive information separation rather than brute-force truncation.
		
			\subsubsection{Dual-Query Abstraction}\label{sec332}

				Following soft disentanglement, we must transform the streams into compact sentiment tokens aligned with textual semantics. To achieve this, we introduce a dual-query mechanism that applies distinct query strategies to aggregate information via cross-attention.
				
				\noindent\textbf{Explicit Sentiment Extraction.} The focus stream $\mathbf{H}_m^{\text{foc}}$ contains salient sentiment shifts. To capture significant variations consistent with human cognition, we utilize a set of predefined text anchors as priors. These anchors describe modality-specific sentiment expressions, such as \textit{"intense body language"} for video and \textit{"rapid tone"} for audio. These descriptions are embedded by an LLM as initialized query vectors $\mathbf{Q}_{\text{foc}}^m$. Subsequently, we perform cross-attention:
				\begin{equation}
					\mathbf{E}_m^{\text{foc}} = \text{CrossAttn}(\mathbf{Q}_{\text{foc}}^m, \mathbf{H}_m^{\text{foc}}, \mathbf{H}_m^{\text{foc}})
				\end{equation}
				Here, $\mathbf{Q}_{\text{foc}}^m$ acts as a semantic probe, retrieving dynamic patterns from the focus stream to match the descriptions. This extracts critical information and explicitly aligns audio-visual signals with the LLM semantic space.
				
				\noindent\textbf{Implicit Ambient Extraction.} The ambient stream $\mathbf{H}_m^{\text{amb}}$ contains unstructured background information like lighting and noise, which is difficult to describe with text. Therefore, we employ randomly initialized learnable parameters as query vectors $\mathbf{Q}_{\text{amb}}^m$. The process is formulated as:
				\begin{equation}
					\mathbf{E}_m^{\text{amb}} = \text{CrossAttn}(\mathbf{Q}_{\text{amb}}^m, \mathbf{H}_m^{\text{amb}}, \mathbf{H}_m^{\text{amb}})
				\end{equation}
				Through end-to-end training, $\mathbf{Q}_{\text{amb}}^m$ adaptively summarizes global background states, providing necessary contextual reference for validating focus stream reliability.

			\subsubsection{Context Calibration}\label{sec333}

				Based on the \textit{Focus-Ambient Calibration} hypothesis, salient features may contain noise and require environmental context to verify them. To achieve this, we utilize the extracted background state $\mathbf{E}_m^{\text{amb}}$ to calibrate the focus sentiment $\mathbf{E}_m^{\text{foc}}$.
				
				Specifically, we design a Confidence Modulation (CFM). This module generates gating coefficients based on background information to amplify or suppress sentiment, controlling reliability. A Feed-Forward Network (FFN) then maps the results to the LLM latent space:
				\begin{equation}
					\begin{aligned}
						\alpha &= \sigma (\text{Linear}(\mathbf{E}_m^{\text{amb}})) \\ 
						\mathbf{E}_m^{\text{mod}} &= \alpha \cdot \mathbf{E}_m^{\text{foc}} \\ 
						\mathbf{Z}_m &= \text{FFN}(\mathbf{E}_m^{\text{foc}} + \mathbf{E}_m^{\text{mod}}) \\ 
					\end{aligned}
				\end{equation}
				where $\alpha$ is the confidence coefficient. The final sentiment token $\mathbf{Z}_m$ integrates structural sentiment shifts with contextual constraints, ensuring high semantic stability and consistency.

			\subsubsection{Unified LLM Reasoning}\label{sec334}
				
				We obtain calibrated sentiment tokens $\mathbf{Z}_v$ and $\mathbf{Z}_a$ after processing. To perform unified reasoning, we concatenate modality features and append a [CLS] \cite{radford2019language} token to build the complete input sequence:
				\begin{equation}
					\mathbf{H}_{\text{in}} = \text{Concat}(\mathbf{Z}_t, \mathbf{Z}_v, \mathbf{Z}_a, [\text{CLS}])
				\end{equation}
				Finally, we perform cross-modal interaction by utilizing the powerful sequence modeling capabilities of LLMs. And we extract the [CLS] token to obtain the sentiment prediction $\hat{y}$.
		
		\subsection{Optimization Objectives}\label{sec34}
		
			To balance prediction accuracy and feature disentanglement, we employ a joint optimization strategy including regression loss and sparsity regularization.
				
			\subsubsection{Task Loss}\label{sec341}
				
				For sentiment prediction on datasets like MOSI/MOSEI, we use Mean Absolute Error (MAE) as the primary loss function:
				\begin{equation}
					\mathcal{L}_{\text{cls}} = \frac{1}{N}\sum_{i=1}^{N} | y_i - \hat{y}_i |
				\end{equation}
				
			\subsubsection{Sparsity Loss}\label{sec342}

				In soft salience disentanglement, the model risks trivial solutions where all $S_t$ are predicted as 1 (keep all) or 0 (drop all). To encourage the model to select truly salient moments, we introduce a target-ratio sparsity loss. This loss constrains the average activation rate near a preset sparsity ratio $k$ (e.g., $k=0.2$ implies only 20\% significant time steps):
				\begin{equation}
					\mathcal{L}_{\text{sp}} = \left| \left( \frac{1}{T}\sum_{t=1}^{T} S_t \right) - k \right|
				\end{equation}
				where $k \in [0,1]$. This explicitly guides the model to focus on the informative time steps. The total optimization objective is defined as:
				\begin{equation}
					\mathcal{L}_{\text{total}} = \mathcal{L}_{\text{cls}} + \lambda \mathcal{L}_{\text{sp}}
				\end{equation}
				where $\lambda$ is a weighted hyperparameter.
				
				\begin{table*}[tp]
					\centering
					\caption{Performance comparison on the MOSI and MOSEI datasets. We compare our method against Traditional and LLM-based baselines. \textbf{Bold}: best result; \underline{Underline}: second best result. $\dag$: reproduced result using public code.}
					\label{tab:main_english_full}
					
					\renewcommand{\arraystretch}{0.95} 
					\setlength{\tabcolsep}{4.0pt} 
					
					\resizebox{\textwidth}{!}{%
						\begin{tabular}{l cccccc c cccccc} 
							\toprule
							\multirow{2}{*}{\textbf{Methods}} & \multicolumn{6}{c}{\textbf{MOSI}} & & \multicolumn{6}{c}{\textbf{MOSEI}} \\
							\cmidrule(lr){2-7} \cmidrule(lr){9-14}
							& \textbf{MAE}$\downarrow$ & \textbf{Corr}$\uparrow$ & \textbf{Acc-7}$\uparrow$ & \textbf{Acc-5}$\uparrow$ & \textbf{Acc-2}$\uparrow$ & \textbf{F1}$\uparrow$ & 
							& \textbf{MAE}$\downarrow$ & \textbf{Corr}$\uparrow$ & \textbf{Acc-7}$\uparrow$ & \textbf{Acc-5}$\uparrow$ & \textbf{Acc-2}$\uparrow$ & \textbf{F1}$\uparrow$ \\
							\midrule
							
							\multicolumn{14}{l}{\textit{\textbf{Traditional Methods}}} \\ 
							\quad TFN \cite{Zadeh2017TensorFN}    & 0.901 & 0.698 & 34.90 & 39.39 & 80.08 & 80.07 && 0.593 & 0.700 & 50.20 & 53.10 & 82.50 & 82.10 \\
							\quad LMF \cite{liu2018efficient}     & 0.917 & 0.695 & 33.20 & 38.13 & 82.50 & 82.40 && 0.623 & 0.677 & 48.00 & 52.90 & 82.00 & 82.10 \\
							\quad Graph-MFN \cite{zadeh2018memory}& 0.956 & 0.649 & 34.64 & 38.63 & 78.35 & 78.35 && 0.575 & 0.713 & 51.37 & 52.69 & 83.48 & 83.43 \\
							\quad MuLT \cite{tsai2019multimodal}  & 0.871 & 0.698 & 40.00 & 42.68 & 83.00 & 82.00 && 0.580 & 0.703 & 51.80 & 54.18 & 82.50 & 82.30 \\
							\quad MISA \cite{Hazarika2020MISAMA}  & 0.777 & 0.778 & 41.37 & 47.08 & 83.54 & 83.58 && 0.558 & 0.752 & 52.05 & 53.63 & 84.67 & 84.66 \\
							\quad MAG-BERT \cite{rahman2020integrating} & 0.727 & 0.781 & 43.62 & - & 84.43 & 84.61 && 0.543 & 0.755 & 52.67 & - & 84.82 & 84.71 \\
							\quad Self-MM \cite{yu2021learning}   & 0.712 & 0.795 & 45.79 & -      & 82.54 & 82.68 && 0.529 & 0.767 & 53.46 & -      & 82.68 & 82.95 \\
							\quad MMIM \cite{han2021improving}    & 0.738 & 0.781 & 45.00 & -      & 85.10 & 85.00 && 0.547 & 0.752 & 53.10 & -      & 85.10 & 85.00 \\
							\quad DMD \cite{li2023decoupled}      & 0.752 & -      & 46.06 & -      & 83.23 & 83.29 && 0.543 & -      & 52.78 & -      & 84.62 & 84.62 \\
							\quad CubeMLP \cite{sun2022cubemlp}   & 0.770 & 0.767 & 45.50 & -      & 85.60 & 85.50 && 0.529 & 0.760 & 54.90 & -      & 85.10 & 84.50 \\
							\quad DLF$^\dag$ \cite{wang2025dlf}          & 0.722 & 0.794 & 46.94 & 53.35 & 85.82 & 85.81 && 0.550 & 0.759 & 52.54 & 54.41 & 84.15 & 84.26 \\
							\quad KAN-MCP$^\dag$ \cite{luo2025towards}   & 0.659 & 0.829 & 46.94 & 53.50 & 86.13 & 86.06 && \underline{0.512} & \underline{0.792} & \underline{54.93} & 56.73 & 86.82 & \underline{86.82} \\
							
							\addlinespace 
							\midrule 
							
							\multicolumn{14}{l}{\textit{\textbf{LLM-based Methods}}} \\
							\quad UniMSE \cite{hu2022unimse}       & 0.691 & 0.809 & \textbf{48.68} & - & 85.85 & 85.83 && 0.523 & 0.773 & 54.39 & - & 85.86 & 85.79 \\
							\quad UniSA-GPT2 \cite{li2023unisa}    & 1.410 & -      & 15.45 & -      & 44.75 & -      && 0.838 & -      & 41.36 & -      & 71.02 & - \\
							\quad UniSA-T5 \cite{li2023unisa}      & 0.900 & -      & 37.46 & -      & 76.82 & -      && 0.546 & -      & 52.50 & -      & 84.22 & - \\
							\quad UniSA-BART \cite{li2023unisa}    & 0.742 & -      & \underline{48.54} & - & 84.11 & - && 0.587 & -      & 50.03 & -      & 84.93 & - \\
							\quad MSE-Qwen-1.8B$^\dag$ \cite{yang2025mse}     & 0.992 & 0.616 & 33.67 & 40.38 & 75.66 & 75.67 && 0.560 & 0.722 & 51.92 & 53.64 & 84.35 & 83.70 \\
							\quad MSE-LLaMA2-7B$^\dag$ \cite{yang2025mse}     & 0.762 & 0.782 & 42.13 & 50.29 & 82.65 & 82.67 && 0.532 & 0.763 & 53.53 & 55.57 & 85.94 & 85.51 \\
							\quad MSE-ChatGLM-6B$^\dag$ \cite{yang2025mse}    & \underline{0.647} & \textbf{0.852} & 46.06 & \underline{54.37} & \underline{88.05} & \underline{88.01} && 0.518 & 0.775 & 54.71 & \underline{56.79} & \underline{86.82} & 86.52 \\
							
							\addlinespace
							\midrule
							
							\multicolumn{14}{l}{\textit{\textbf{Proposed Method}}} \\ 
							\quad Vanilla SentiLLM   & 0.823 & 0.765 & 36.73 & 43.73 & 83.99 & 83.79 && 0.529 & 0.782 & 53.17 & 54.82 & 86.71 & 86.71 \\
							
							\rowcolor{graybg} 
							\quad \textbf{SentiLLM (Ours)}        & \textbf{0.639} & \underline{0.846} & \textbf{48.68} & \textbf{56.71} & \textbf{89.02} & \textbf{88.99} && \textbf{0.508} & \textbf{0.799} & \textbf{55.29} & \textbf{56.84} & \textbf{87.29} & \textbf{87.29} \\
							\bottomrule
						\end{tabular}%
					}
				\end{table*}

	\section{Experiments}\label{sec4}
	
		\subsection{Datasets and Evaluation Metrics}\label{sec41}
		
			\noindent\textbf{Datasets.} We evaluate our approach on four datasets: CMU-MOSI \cite{Zadeh2016MOSIMC}, CMU-MOSEI \cite{zadeh2018multimodal}, CH-SIMS \cite{yu2020ch}, and CH-SIMS v2 \cite{liu2022make}. Detailed dataset statistics are provided in the \textit{Supplementary Material}.
			
			\noindent\textbf{Evaluation Metrics.} To ensure fair comparison, we align with prior works \cite{wang2025dlf,yu2020ch,liu2022make}. For MOSI and MOSEI, we report 7-class accuracy (Acc-7), 5-class accuracy (Acc-5), binary accuracy (Acc-2), F1 score (F1), mean absolute error (MAE) and Pearson correlation (Corr). For CH-SIMS and CH-SIMS v2, we utilize Acc-5, Acc-3, Acc-2, F1, MAE, and Corr.
		
		\subsection{Baseline Methods}\label{sec42}
		
			We compare our method against state-of-the-art models, categorized into: 
			\begin{enumerate}
				\item \textbf{Traditional Methods}: TFN \cite{Zadeh2017TensorFN}, LMF \cite{liu2018efficient}, Graph-MFN \cite{zadeh2018multimodal}, MuLT \cite{tsai2019multimodal}, MISA \cite{Hazarika2020MISAMA}, MAG-BERT \cite{rahman2020integrating}, Self-MM \cite{yu2021learning}, MMIM \cite{han2021improving}, AV-MC \cite{liu2022make}, DMD \cite{li2023decoupled}, CubeMLP \cite{sun2022cubemlp}, DLF \cite{wang2025dlf}, and KAN-MCP \cite{luo2025towards}; 
				
				\item \textbf{LLM-based Methods}: UniMSE \cite{hu2022unimse}, UniSA \cite{li2023unisa} and MSE-Adapter \cite{yang2025mse}; 
				
				\item \textbf{Vanilla SentiLLM (Baseline)}: A variant without our proposed module, using only linear layers to project LLM inputs. 
			\end{enumerate}
		
		\subsection{Implementation Details}\label{sec43}
		
			\noindent\textbf{Feature Extraction.} For MOSI and MOSEI, we follow \cite{wang2025dlf} to obtain raw text descriptions, video, and audio features. For CH-SIMS and CH-SIMS v2, we utilize data files provided by \cite{yu2020ch} and \cite{liu2022make}, respectively.
			
			\noindent\textbf{Settings.} We conduct all experiments on NVIDIA RTX 5090 32GB GPUs. For LLM, we use Qwen2.5-1.5B \cite{Yang2024Qwen25TR}, initializing its parameters from the official release\footnote{\url{https://huggingface.co/Qwen/Qwen2.5-1.5B}}. The input dimension $D$ is set to 256. The hyperparameters are adjusted based on the dataset’s scale, as detailed in \textit{Supplementary Material}. Notably, SentiLLM has 1549M parameters (Qwen2.5-1.5B: 1544M, Proposed Module: 5M), \textit{with only 0.32\% trainable}, substantially fewer than mainstream approaches.
			
			\noindent\textbf{More Experimental Results.} We defer detailed analyses on the impact of different modalities, the roles of soft salience disentanglement, the effectiveness of sparsity loss, and additional visualizations to the \textit{Supplement Material}.
			
		\subsection{Comparison with State-of-the-arts}\label{sec44}
		
			\begin{table*}[t]
				\centering
				\caption{Performance comparison on the CH-SIMS and CH-SIMS v2 datasets. We compare our method against Traditional and LLM-based baselines. \textbf{Bold}: best result; \underline{Underline}: second best result. $\dag$: reproduced result using public code.}
				\label{tab:main_chinese_full}
				
				\renewcommand{\arraystretch}{0.90} 
				\setlength{\tabcolsep}{4.0pt} 
				
				\resizebox{\textwidth}{!}{%
					\begin{tabular}{l cccccc c cccccc}
						\toprule
						\multirow{2}{*}{\textbf{Methods}} & \multicolumn{6}{c}{\textbf{CH-SIMS}} & & \multicolumn{6}{c}{\textbf{CH-SIMS v2}} \\
						\cmidrule(lr){2-7} \cmidrule(lr){9-14}
						& \textbf{MAE}$\downarrow$ & \textbf{Corr}$\uparrow$ & \textbf{Acc-5}$\uparrow$ & \textbf{Acc-3}$\uparrow$ & \textbf{Acc-2}$\uparrow$ & \textbf{F1}$\uparrow$ & 
						& \textbf{MAE}$\downarrow$ & \textbf{Corr}$\uparrow$ & \textbf{Acc-5}$\uparrow$ & \textbf{Acc-3}$\uparrow$ & \textbf{Acc-2}$\uparrow$ & \textbf{F1}$\uparrow$ \\
						\midrule
						
						\multicolumn{14}{l}{\textit{\textbf{Traditional Methods}}} \\
						\quad TFN \cite{Zadeh2017TensorFN}     & 0.437 & 0.582 & \na & \na & 77.10 & 76.90 && 0.322 & 0.662 & 53.30 & 70.90 & 78.10 & 78.10 \\
						\quad LMF \cite{liu2018efficient}      & 0.438 & 0.578 & \na & \na & 77.40 & 77.40 && 0.327 & 0.651 & 51.60 & 70.00 & 77.80 & 77.80 \\
						\quad MuLT \cite{tsai2019multimodal}   & 0.442 & 0.581 & 40.00 & 65.70 & 78.20 & 78.50 && 0.300 & \textbf{0.738} & \underline{54.60} & \underline{74.20} & \underline{80.80} & \underline{80.70} \\
						\quad MISA \cite{Hazarika2020MISAMA}   & 0.447 & 0.563 & \na & \na & 76.50 & 76.60 && 0.342 & 0.671 & 47.50 & 68.90 & 78.20 & 78.30 \\
						\quad MAG-BERT$^\dag$ \cite{rahman2020integrating} & 0.457 & 0.559 & 37.20 & 65.65 & 75.49 & 75.94 && 0.338 & 0.651 & 49.32 & 69.54 & 76.50 & 76.63 \\
						\quad Self-MM \cite{yu2021learning}    & 0.411 & 0.601 & 43.10 & 66.10 & 78.60 & 78.60 && 0.315 & 0.691 & 53.50 & 72.70 & 78.70 & 78.60 \\
						\quad MMIM \cite{han2021improving}     & \underline{0.422} & 0.597 & 42.00 & 65.50 & 78.30 & 78.20 && 0.339 & 0.641 & 50.50 & 70.40 & 77.80 & 77.80 \\
						\quad AV-MC$^\dag$ \cite{liu2022make}  & \textbf{0.378} & \textbf{0.675} & \textbf{45.08} & \textbf{70.02} & \underline{80.31} & \textbf{80.49} && \underline{0.298} & \underline{0.733} & 53.58 & 72.44 & 80.37 & 80.25 \\
						\quad DLF$^\dag$ \cite{wang2025dlf}    & 0.547 & 0.222 & 27.57 & 51.64 & 69.80 & 62.78 && 0.435 & 0.430 & 35.20 & 52.13 & 67.70 & 67.40 \\
						\quad KAN-MCP$^\dag$ \cite{luo2025towards} & 0.493 & 0.452 & 35.67 & 59.96 & 72.87 & 73.07 && 0.392 & 0.525 & 40.91 & 60.64 & 71.37 & 71.36 \\
						
						\addlinespace
						\midrule
						
						\multicolumn{14}{l}{\textit{\textbf{LLM-based Methods}}} \\
						\quad MSE-Qwen-1.8B$^\dag$ \cite{yang2025mse}   & \na & \na & \na & \na & \na & \na && 0.318 & 0.668 & \textbf{54.74} & 72.63 & 79.59 & 79.64 \\
						\quad MSE-LLaMA2-7B$^\dag$ \cite{yang2025mse}  & \na & \na & \na & \na & \na & \na && 0.393 & 0.537 & 47.78 & 69.15 & 75.73 & 75.83 \\
						\quad MSE-ChatGLM-6B$^\dag$ \cite{yang2025mse} & \na & \na & \na & \na & \na & \na && \textbf{0.290} & 0.730 & 54.06 & \textbf{76.02} & \textbf{81.91} & \textbf{81.87} \\
						
						\addlinespace
						\midrule
						
						\multicolumn{14}{l}{\textit{\textbf{Proposed Method}}} \\
						\quad Vanilla SentiLLM        & 0.478 & 0.508 & 35.01 & 60.18 & 74.62 & 74.40 && 0.368 & 0.630 & 43.23 & 65.47 & 76.11 & 76.19 \\
						
						\rowcolor{graybg}
						\quad \textbf{SentiLLM (Ours)} & \underline{0.422} & \underline{0.616} & \underline{44.20} & \underline{67.18} & \textbf{80.96} & \underline{80.22} && 0.329 & 0.689 & 48.55 & 71.18 & 79.88 & 79.91 \\
						\bottomrule
					\end{tabular}%
				}
			\end{table*}
					
			\noindent\textbf{MOSI \& MOSEI.} As shown in Table~\ref{tab:main_english_full}, our proposed method achieves state-of-the-art performance on both MOSI and MOSEI datasets, significantly outperforming traditional baselines and recent LLM-based approaches. On MOSI, our method achieves the best Acc-2 of 89.02\% and Acc-7 of 48.68\%, surpassing the strongest competitor, MSE-ChatGLM3-6B, by 0.97\% in accuracy. Notably, our method utilizes a much smaller backbone compared to LLaMA2-7B and ChatGLM3-6B, yet it yields superior results. This efficiency is attributed to our dual-stream mechanism, which extract semantically aligned tokens rather than relying solely on the LLM's inherent capacity. Furthermore, the significant margin between our method and the Vanilla SentiLLM highlights the critical role of our structural abstraction mechanism in bridging the modality gap, preventing the LLM from being overwhelmed by noisy, unaligned multimodal signals.
			
			\noindent\textbf{CH-SIMS \& CH-SIMS v2.} We further evaluate the generalization capability of our method on Chinese sentiment datasets, as shown in Table~\ref{tab:main_chinese_full}. On CH-SIMS, our method establishes a new state-of-the-art Acc-2 of 80.96\%, outperforming the robust AV-MC baseline. On the larger CH-SIMS v2 dataset, while MSE-ChatGLM3-6B achieves the top performance due to its larger parameter size and native Chinese pre-training, our method remains highly competitive, achieving 79.88\% Acc-2 and surpassing MSE-LLaMA2-7B (+4.15\%) and MSE-Qwen-1.8B (+0.29\%). Crucially, our approach demonstrates consistent improvements over the Vanilla SentiLLM across both datasets, confirming that our concept of learning structural abstraction is linguistically universal. This suggests that our method effectively captures universal salient sentiment shifts, independent of the specific language, making it a robust solution for cross-lingual multimodal sentiment analysis.

		\subsection{Ablation Studies}\label{sec45}

			\begin{table}[tp]
				\centering
				\caption{Ablation study on the contribution of different model components on the MOSI dataset.}
				\label{tab:ablation}
				
				\renewcommand{\arraystretch}{0.90}
				\setlength{\tabcolsep}{4pt}
				
				\resizebox{\columnwidth}{!}{
					\begin{tabular}{lcccccc}
						\toprule
						\textbf{Model Configuration} & \textbf{Step} & \textbf{MAE} $\downarrow$ & \textbf{Corr} $\uparrow$ & \textbf{Acc-7} $\uparrow$ & \textbf{Acc-2} $\uparrow$ & \textbf{F1} $\uparrow$ \\
						\midrule
						\textbf{SentiLLM (Ours)} & 0 & \textbf{0.640} & \underline{0.846} & \textbf{48.40} & \textbf{89.02} & \textbf{88.99} \\
						
						\midrule
						\quad \textit{w/o} Context Calibration & 1 & \underline{0.643} & \textbf{0.849} & \underline{46.21} & \underline{88.87} & \underline{88.88} \\
						\quad \textit{w/o} Focus Stream & 2 & 0.682 & 0.832 & 43.59 & 87.65 & 87.61 \\
						\quad \textit{w/o} Ambient Stream & 2 & 0.676 & 0.829 & 44.61 & 87.80 & 87.74 \\
						\quad \textit{w/o} Dual Stream & 3 & 0.699 & 0.812 & 44.17 & 85.67 & 85.66 \\
						
						\midrule
						Vanilla SentiLLM & 4 & 0.823 & 0.765 & 36.73 & 83.99 & 83.79 \\
						\bottomrule
					\end{tabular}
				}
			\end{table}
		
			\begin{table}[tp]
				\centering
				\caption{Impact of different query initialization strategies on the MOSI dataset.}
				\label{tab:query_impact}
				
				\renewcommand{\arraystretch}{0.90}
				\setlength{\tabcolsep}{4pt}
				
				\resizebox{\columnwidth}{!}{%
					\begin{tabular}{lcccccc}
						\toprule
						\textbf{Prompt Strategy} & \textbf{MAE} $\downarrow$ & \textbf{Corr} $\uparrow$ & \textbf{Acc-7} $\uparrow$ & \textbf{Acc-5} $\uparrow$ & \textbf{Acc-2} $\uparrow$ & \textbf{F1} $\uparrow$ \\
						\midrule
						\textbf{Semantic Sentences} & \textbf{0.640} & \textbf{0.846} & \textbf{48.40} & \textbf{56.71} & \textbf{89.02} & \textbf{88.99} \\
						Simplified Keywords & \underline{0.644} & \underline{0.840} & \underline{47.52} & \underline{54.96} & \underline{87.35} & \underline{87.31} \\
						Random Vectors & 0.684 & 0.828 & 43.44 & 50.58 & 86.74 & 86.71 \\
						\bottomrule
					\end{tabular}%
				}
			\end{table}
		
			\noindent\textbf{Effectiveness of Model Components.} The contribution of each component is reported in Table~\ref{tab:ablation}. The significant performance drop is observed when removing the ambient stream or the dual-stream structure. This highlights the importance of contextual background alongside salient sentiment shifts. Additionally, the contextual calibration is crucial for fine-grained improvements, confirming the necessity of calibrating salient sentiment shifts.
		
			\noindent\textbf{Impact of Query Initialization.} The results in Table~\ref{tab:query_impact} examine the impact of query initialization. This demonstrates that our detailed semantic sentences consistently outperform simplified keywords (e.g., positive/negative) or random vectors. The performance gap confirms that leveraging the pre-aligned LLM semantic space is crucial for guiding the attention mechanism to capture subtle, structurally meaningful affective cues.
			
			\begin{table*}[t]
				\centering
				\caption{Generalization analysis across different LLM families on MOSI and CH-SIMS datasets. ``$\Delta$'' denotes the performance gap between Ours and Baseline. \textcolor{goodgreen}{Green} indicates improvement, while \textcolor{badred}{Red} indicates degradation.}
				\label{tab:llm_family_generalization}
				
				\renewcommand{\arraystretch}{0.90} 
				\setlength{\tabcolsep}{4pt} 
				
				\resizebox{0.95\textwidth}{!}{%
					\begin{tabular}{llcccccccccc}
						\toprule
						\multirow{2}{*}{\textbf{LLM Backbone}} & \multirow{2}{*}{\textbf{Setting}} & \multicolumn{5}{c}{\textbf{MOSI}} & \multicolumn{5}{c}{\textbf{CH-SIMS}} \\
						\cmidrule(lr){3-7} \cmidrule(lr){8-12}
						& & \textbf{MAE} $\downarrow$ & \textbf{Corr} $\uparrow$ & \textbf{Acc-7} $\uparrow$ & \textbf{Acc-2} $\uparrow$ & \textbf{F1} $\uparrow$ & \textbf{MAE} $\downarrow$ & \textbf{Corr} $\uparrow$ & \textbf{Acc-5} $\uparrow$ & \textbf{Acc-2} $\uparrow$ & \textbf{F1} $\uparrow$ \\
						\midrule
						
						\multirow{3}{*}{\textbf{Qwen2.5-1.5B}} 
						& Baseline & 0.823 & 0.765 & 36.73 & 83.99 & 83.79 & 0.478 & 0.508 & 35.01 & 74.62 & 74.40 \\
						& \textbf{Ours} & \best{0.640} & \best{0.846} & \best{48.40} & \best{89.02} & \best{88.99} & \best{0.422} & \best{0.616} & \best{44.20} & \best{80.96} & \best{80.22} \\
						& \textit{$\Delta$} & \cneg{-0.183} & \cpos{0.081} & \cpos{11.67} & \cpos{5.03} & \cpos{5.20} & \cneg{-0.056} & \cpos{0.108} & \cpos{9.19} & \cpos{6.34} & \cpos{5.82} \\
						\midrule
						
						\multirow{3}{*}{\textbf{InternLM2.5-1.8B}} 
						& Baseline & 0.958 & 0.662 & 31.78 & 78.66 & 78.56 & \best{0.431} & 0.599 & 41.14 & 79.87 & 79.29 \\
						& \textbf{Ours} & \best{0.703} & \best{0.825} & \best{44.31} & \best{85.52} & \best{85.27} & 0.433 & \best{0.600} & \best{41.94} & \best{80.31} & \best{80.58} \\
						& \textit{$\Delta$} & \cneg{-0.255} & \cpos{0.163} & \cpos{12.53} & \cpos{6.86} & \cpos{6.71} & \cbadneg{0.002} & \cpos{0.001} & \cpos{0.80} & \cpos{0.44} & \cpos{1.29} \\
						\midrule
						
						\multirow{3}{*}{\textbf{Gemma-2-2B-it}} 
						& Baseline & 0.725 & 0.805 & 43.00 & 85.98 & 85.98 & 0.433 & 0.610 & 42.23 & \best{78.12} & 76.79 \\
						& \textbf{Ours} & \best{0.599} & \best{0.865} & \best{49.56} & \best{89.18} & \best{89.12} & \best{0.428} & \best{0.612} & \best{42.89} & 77.90 & \best{77.04} \\
						& \textit{$\Delta$} & \cneg{-0.126} & \cpos{0.060} & \cpos{6.56} & \cpos{3.20} & \cpos{3.14} & \cneg{-0.005} & \cpos{0.002} & \cpos{0.66} & \cbadpos{-0.22} & \cpos{0.25} \\
						\midrule
						
						\multirow{3}{*}{\textbf{Llama-3.2-3B}} 
						& Baseline & 0.793 & 0.769 & 39.65 & 84.45 & 84.43 & 0.569 & 0.503 & 36.76 & 74.84 & 69.47 \\
						& \textbf{Ours} & \best{0.669} & \best{0.832} & \best{44.75} & \best{87.20} & \best{87.18} & \best{0.493} & \best{0.554} & \best{37.64} & \best{77.24} & \best{74.46} \\
						& \textit{$\Delta$} & \cneg{-0.124} & \cpos{0.063} & \cpos{5.10} & \cpos{2.75} & \cpos{2.75} & \cneg{-0.076} & \cpos{0.051} & \cpos{0.88} & \cpos{2.40} & \cpos{4.99} \\
						\bottomrule
					\end{tabular}%
				}
			\end{table*}
			
\begin{figure}[t]
				\centering
				\includegraphics[width=0.85\columnwidth]{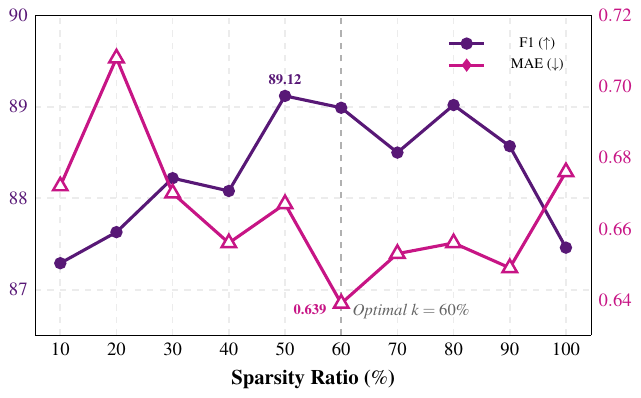}
				\caption{Ablation study on the impact of sparsity ratio $k$ on the MOSI dataset.}
				\label{fig:exp:topk}
			\end{figure}

			\noindent\textbf{Analysis of sparsity ratio $k$.} As illustrated in Figure~\ref{fig:exp:topk}, model performance exhibits a distinct pattern regarding sparsity ratio. Low sparsity causes severe information loss by discarding salient moments, whereas high sparsity introduces excessive noise, interfering with LLM reasoning. An optimal sparsity ratio achieves the best trade-off between retaining sentiment shifts and background states, validating the effectiveness of soft salience disentanglement.
			
\begin{figure}[t]
				\centering
				\includegraphics[width=0.85\columnwidth]{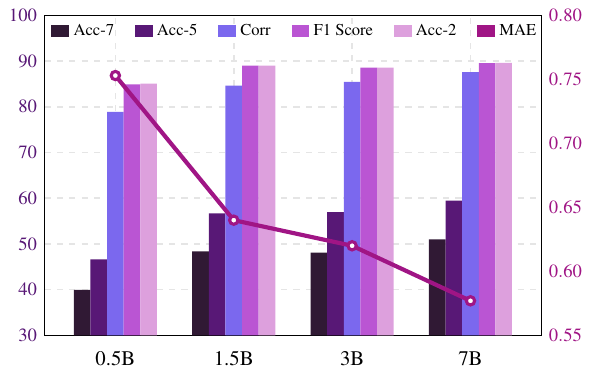}
				\caption{Scalability analysis with different Qwen2.5 sizes on the MOSI dataset.}
				\label{fig:exp:llm_size}
			\end{figure}

			\noindent\textbf{LLMs Scalability.} We investigate scalability by varying the LLM size from 0.5B to 7B parameters. As shown in Figure~\ref{fig:exp:llm_size}, our method exhibits strong scalability properties. Although Acc-2 fluctuates slightly at 3B, the strict monotonic decrease in MAE indicates that larger models provide more precise prediction scores. This suggests that our method effectively leverages the capabilities of larger LLMs.
			
			\noindent\textbf{Generalization to Different LLMs.} As presented in Table~\ref{tab:llm_family_generalization}, to demonstrates generalization capabilities, we evaluate four distinct LLMs: Qwen \cite{Yang2024Qwen25TR}, InternLM \cite{cai2024internlm2}, Gemma \cite{team2024gemma}, and Llama3 \cite{dubey2024llama}. On the English-dominant MOSI, our method yields consistent gains, notably boosting the Chinese-optimized InternLM by +6.86\% (Acc-2). On the Chinese CH-SIMS, results reflect linguistic characteristics: the bilingual Qwen benefits most significantly (+6.34\%), demonstrating perfect synergy. While the English-centric Gemma faces a slight accuracy trade-off due to cross-lingual alignment gaps, its improved F1 score confirms structural robustness. Even against InternLM's exceptionally high baseline, our method maintains performance gains, verifying that SentiLLM provides a universal, semantic-aligned structural abstraction, regardless of the underlying pre-training distribution.
			
		\subsection{Extended Analysis}\label{sec46}

			\begin{figure*}[t]
				\centering
				\setlength{\tabcolsep}{1pt} 
				
				\resizebox{0.85\linewidth}{!}{%
					\begin{minipage}{\linewidth}
				
						\hspace{1.5em} 
						\makebox[0.23\linewidth]{\small \textbf{MOSI}} \hfill
						\makebox[0.23\linewidth]{\small \textbf{MOSEI}} \hfill
						\makebox[0.23\linewidth]{\small \textbf{CH-SIMS}} \hfill
						\makebox[0.23\linewidth]{\small \textbf{CH-SIMS v2}}
		
						\rotatebox{90}{\makebox[0.23\linewidth][c]{\small \textbf{Baseline}}}%
						\hspace{0.5em}%
						\begin{subfigure}[b]{0.23\linewidth}
							\includegraphics[width=\textwidth]{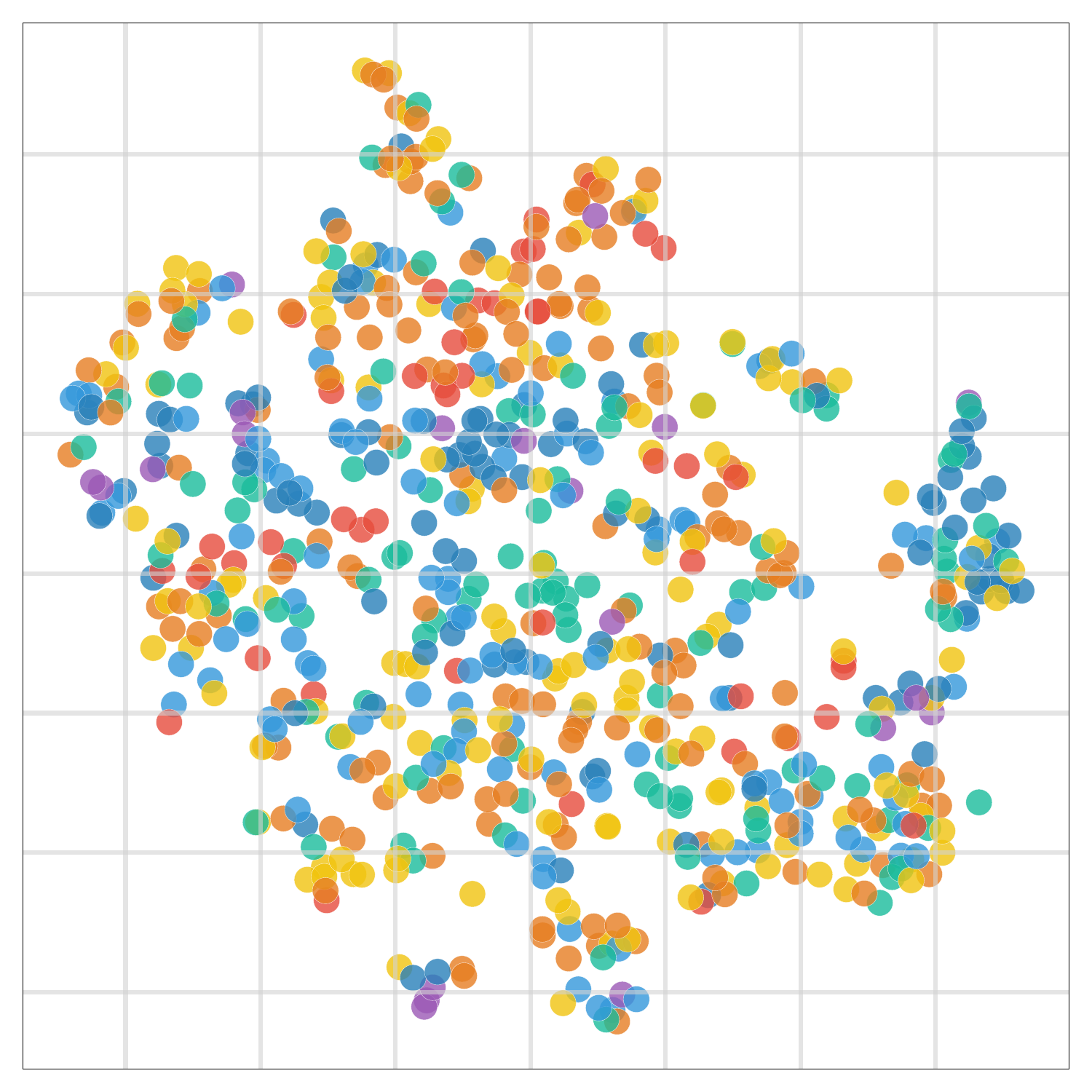}
						\end{subfigure}\hfill
						\begin{subfigure}[b]{0.23\linewidth}
							\includegraphics[width=\textwidth]{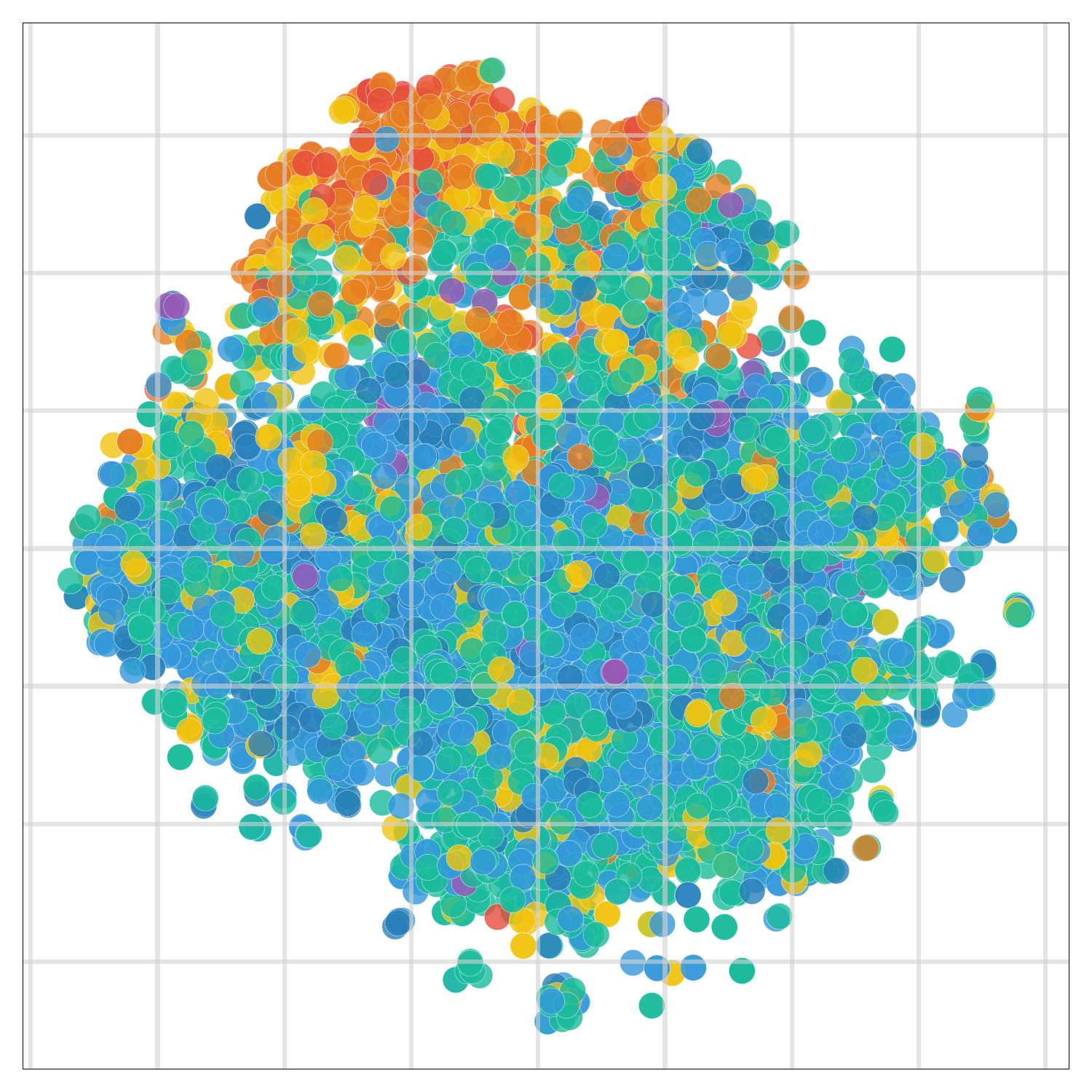}
						\end{subfigure}\hfill
						\begin{subfigure}[b]{0.23\linewidth}
							\includegraphics[width=\textwidth]{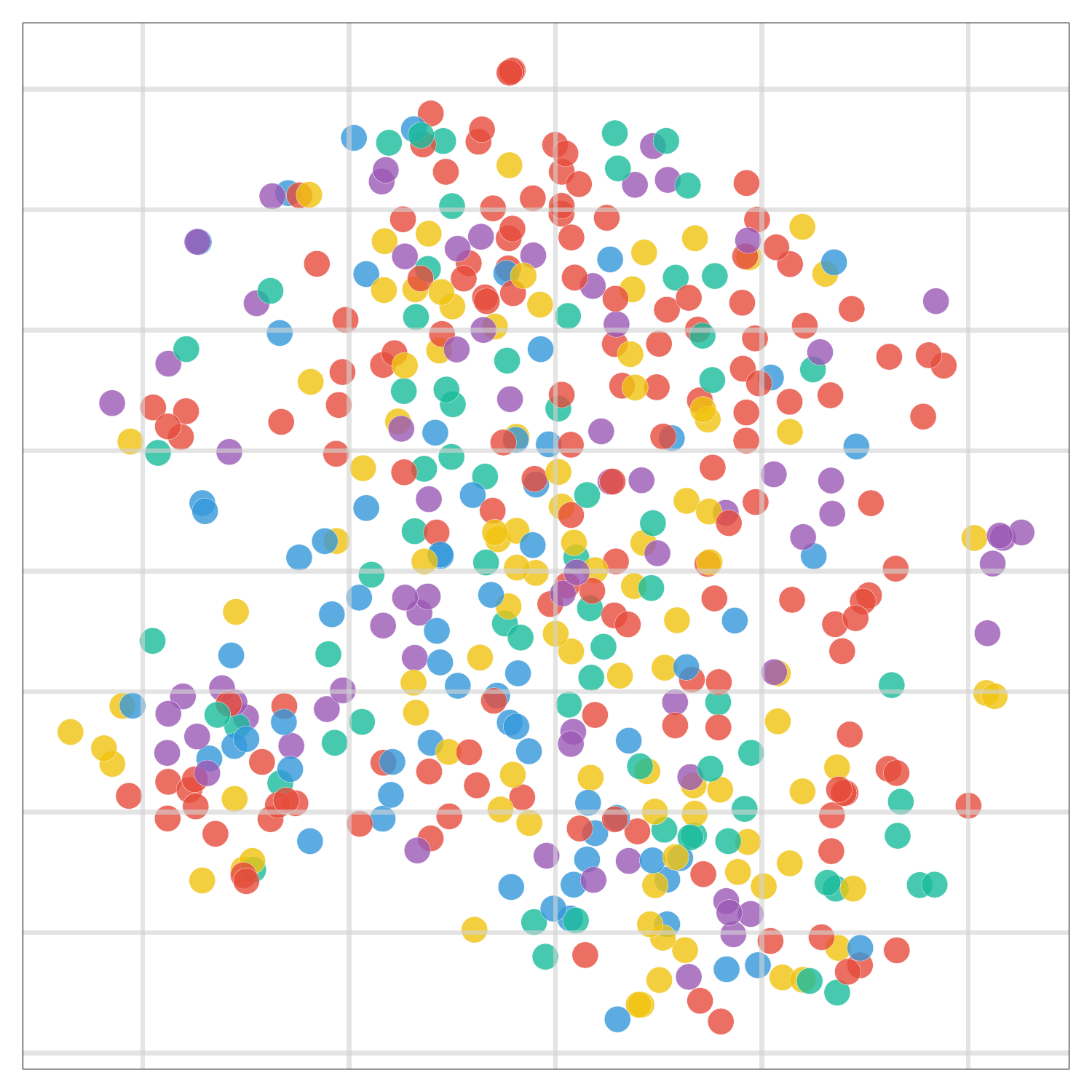}
						\end{subfigure}\hfill
						\begin{subfigure}[b]{0.23\linewidth}
							\includegraphics[width=\textwidth]{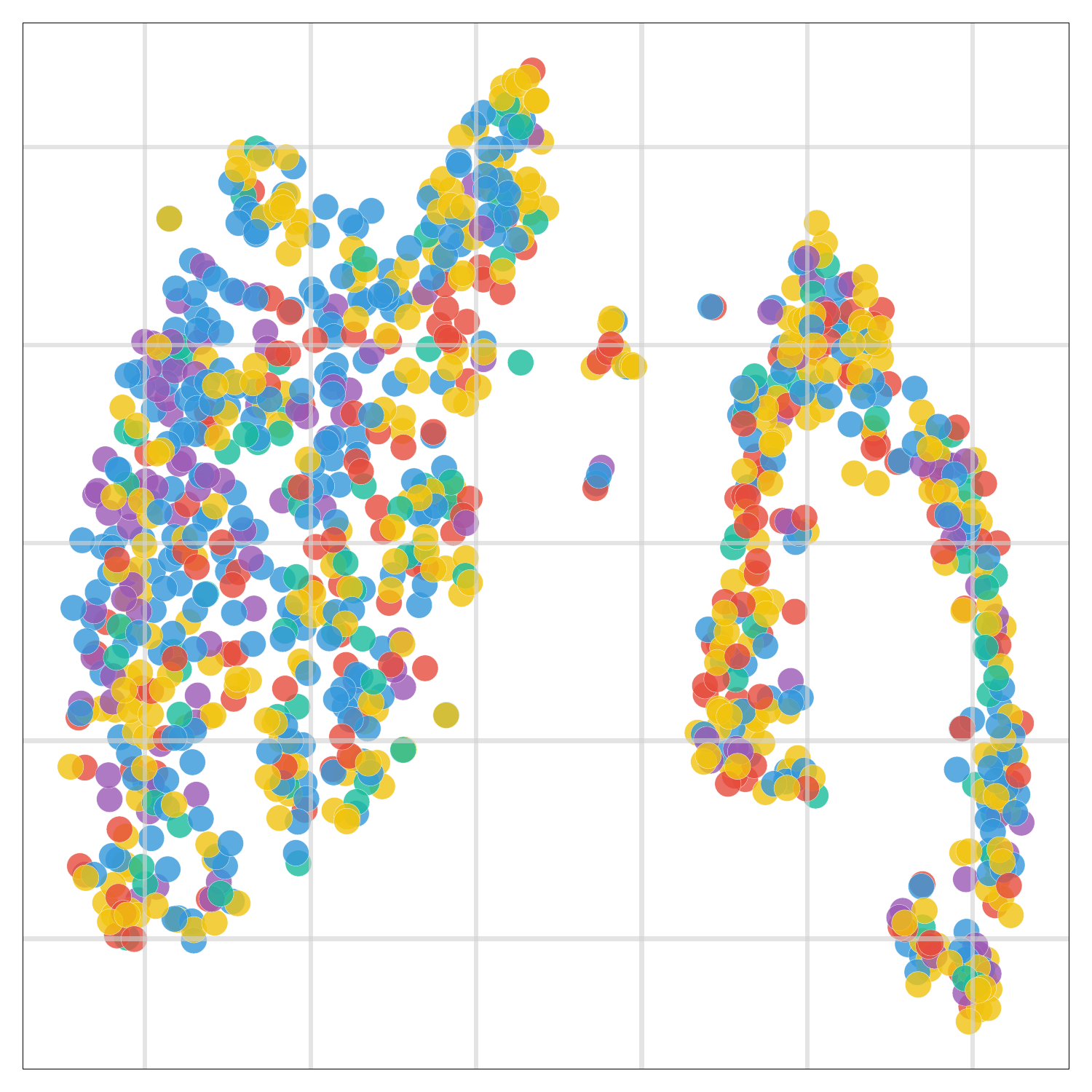}
						\end{subfigure}
		
						\rotatebox{90}{\makebox[0.23\linewidth][c]{\small \textbf{Ours}}}%
						\hspace{0.5em}%
						\begin{subfigure}[b]{0.23\linewidth}
							\includegraphics[width=\textwidth]{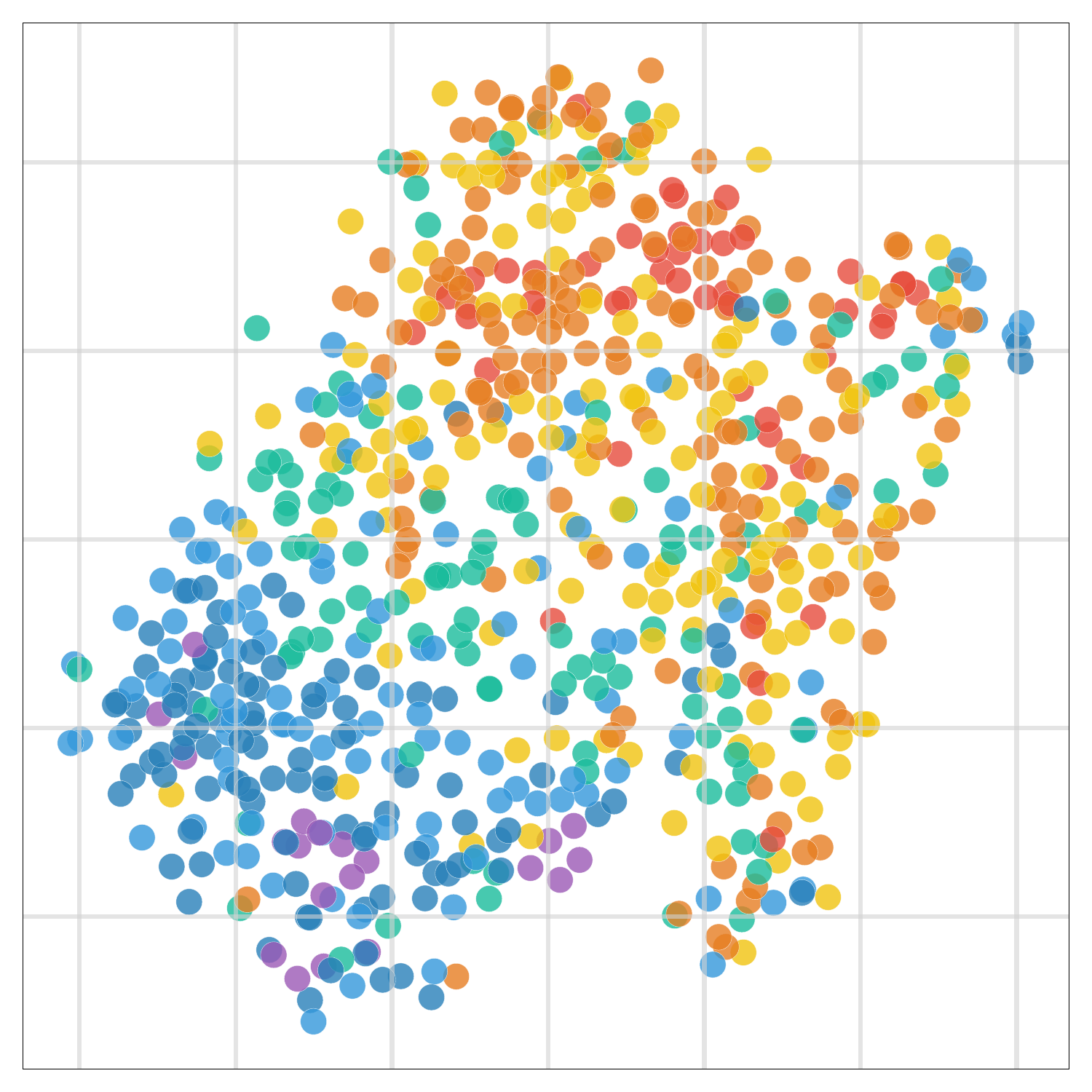}
						\end{subfigure}\hfill
						\begin{subfigure}[b]{0.23\linewidth}
							\includegraphics[width=\textwidth]{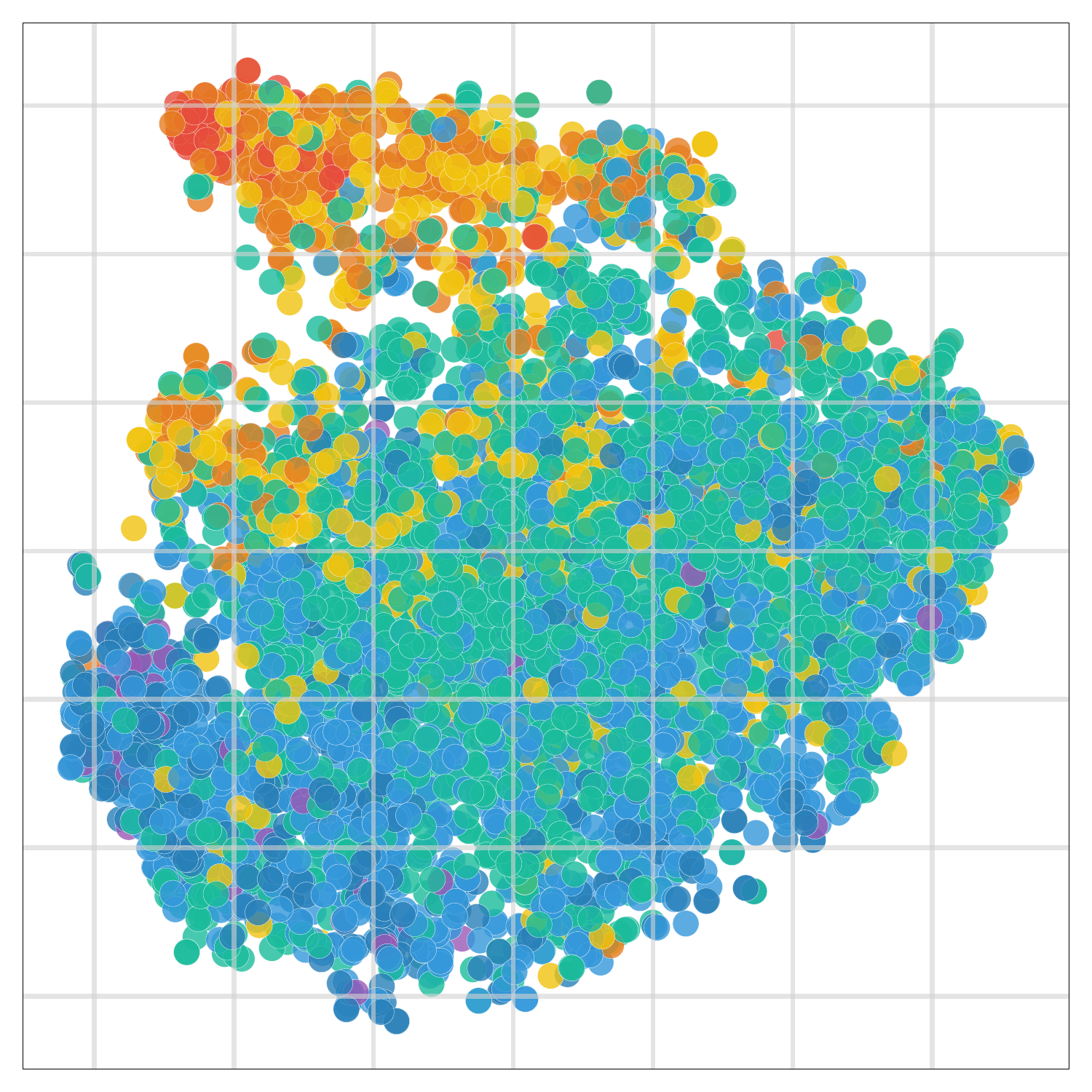}
						\end{subfigure}\hfill
						\begin{subfigure}[b]{0.23\linewidth}
							\includegraphics[width=\textwidth]{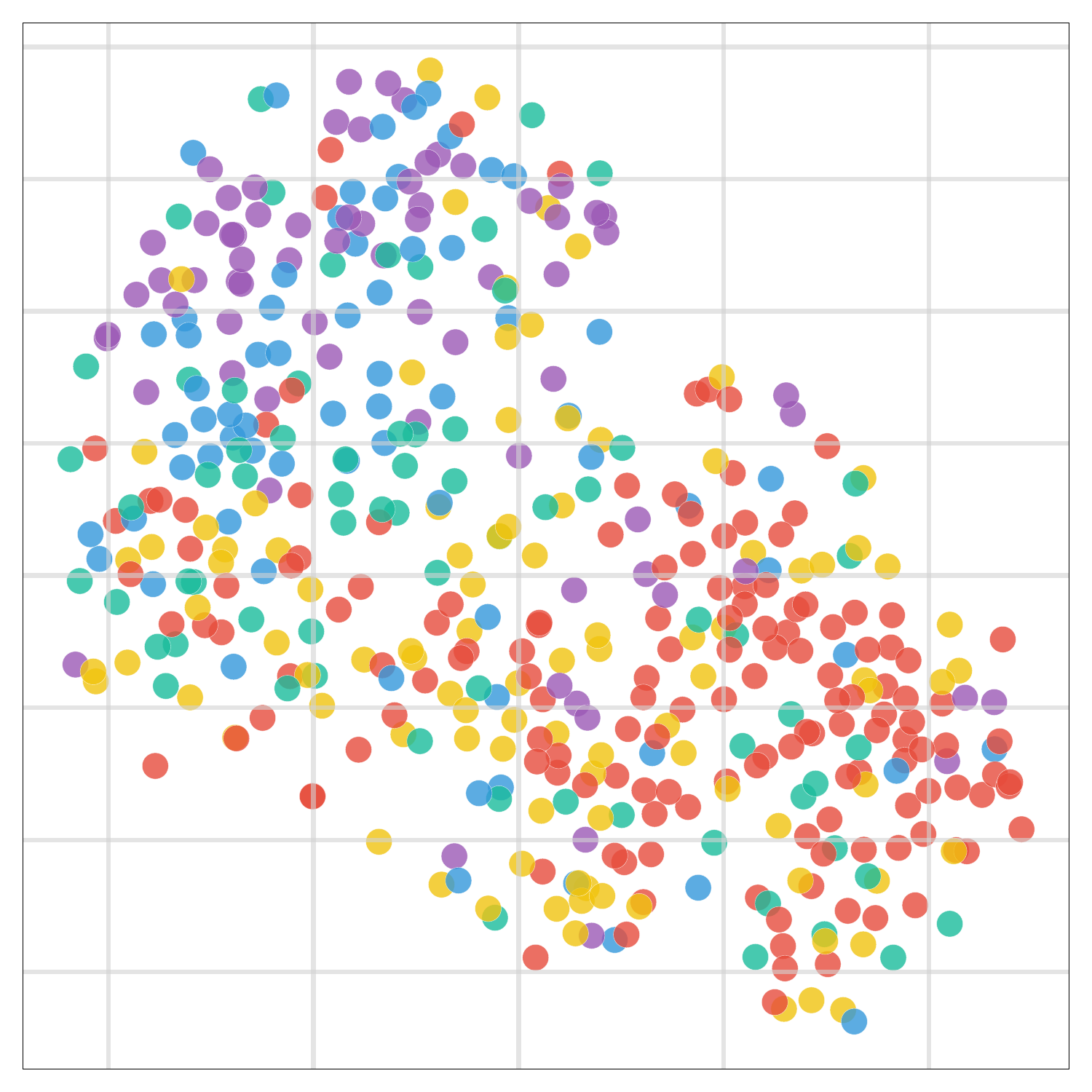}
						\end{subfigure}\hfill
						\begin{subfigure}[b]{0.23\linewidth}
							\includegraphics[width=\textwidth]{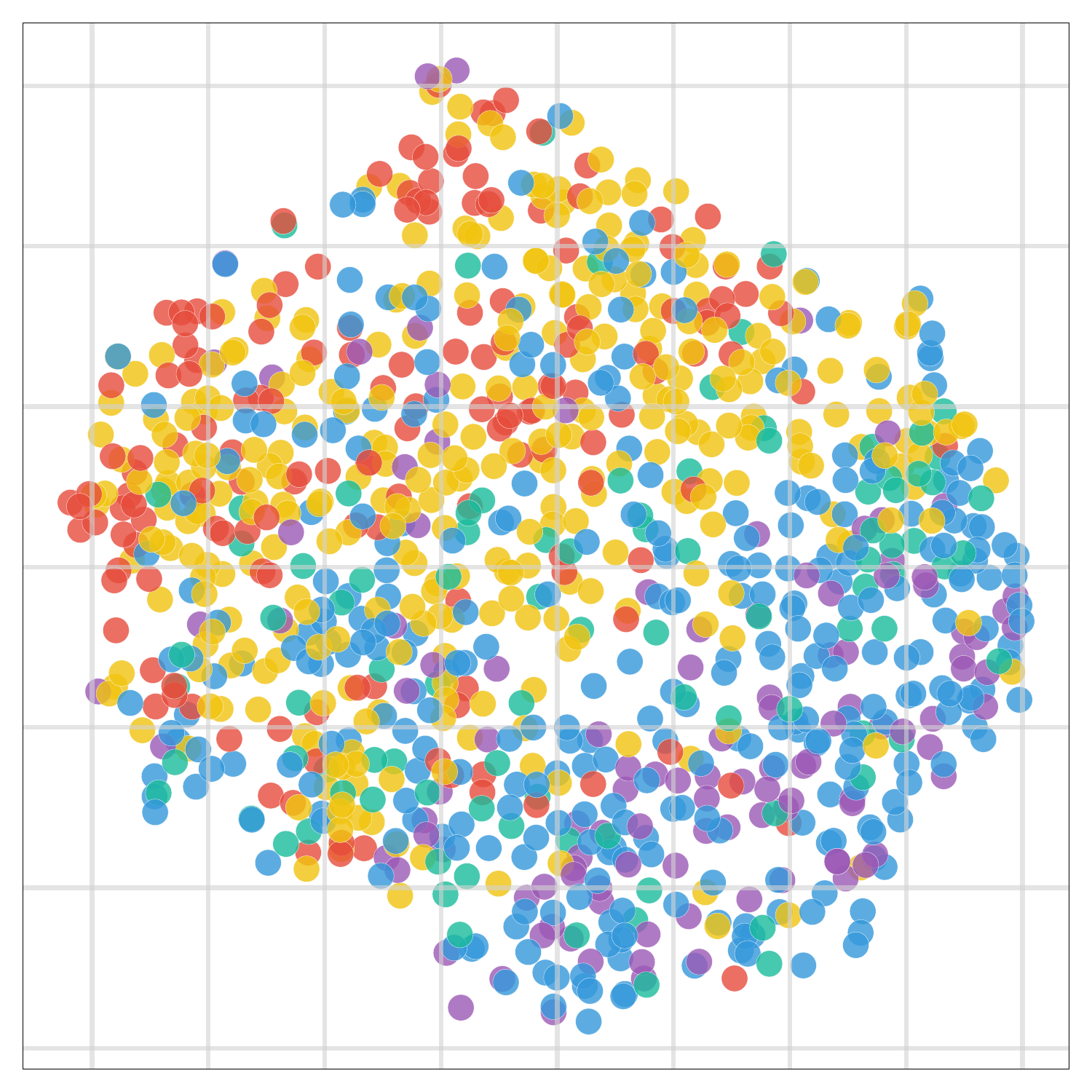}
						\end{subfigure}
						
						\begin{center}
							{\footnotesize
								\centering
								\sentimentdot{sneg3} Highly Negative \quad
								\sentimentdot{sneg2} Negative \quad
								\sentimentdot{sneg1} Weak Negative \quad
								\sentimentdot{sneu0} Neutral \quad 
								\sentimentdot{spos1} Weak Positive \quad
								\sentimentdot{spos2} Positive \quad
								\sentimentdot{spos3} Highly Positive
							}
						\end{center}
					\end{minipage}
				}
				
				\caption{Visualization of learned representation spaces. Rows indicate the methods (Baseline vs. Ours) and columns represent different datasets. The color legend represents sentiment intensity classes.}
				\label{fig:tsne_visualization_8grid}
			\end{figure*}
			
			\begin{figure}[t]
				\centering
				\resizebox{0.85\columnwidth}{!}{ 
					\begin{tabular}{c c c} 
						& \small \textbf{Video Modal} & \small \textbf{Audio Modal} \\
						\rotatebox[origin=c]{90}{\small \textbf{Focus Stream}} & 
						\begin{subfigure}[c]{0.44\columnwidth}
							\includegraphics[width=\linewidth]{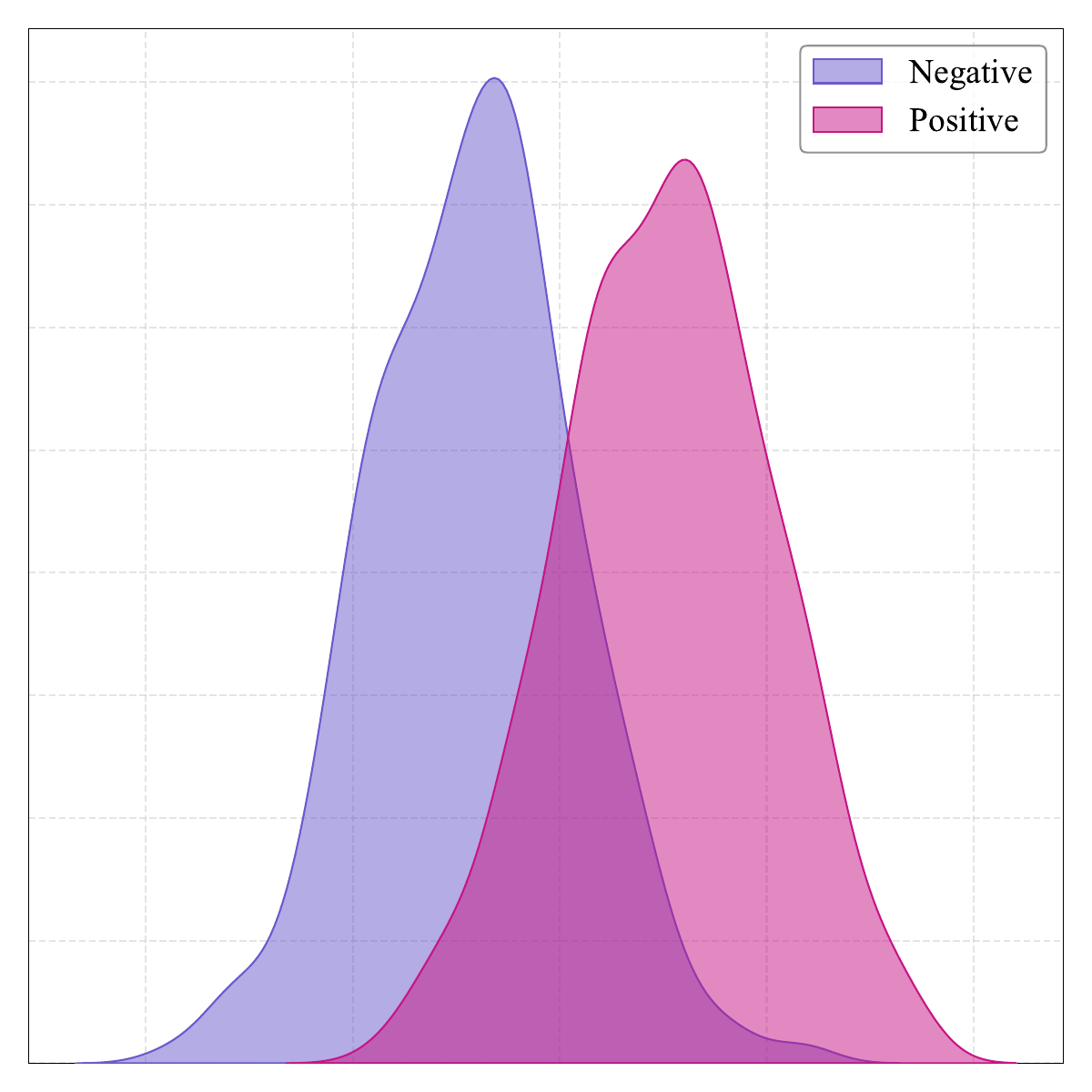}
						\end{subfigure} & 
						\begin{subfigure}[c]{0.44\columnwidth}
							\includegraphics[width=\linewidth]{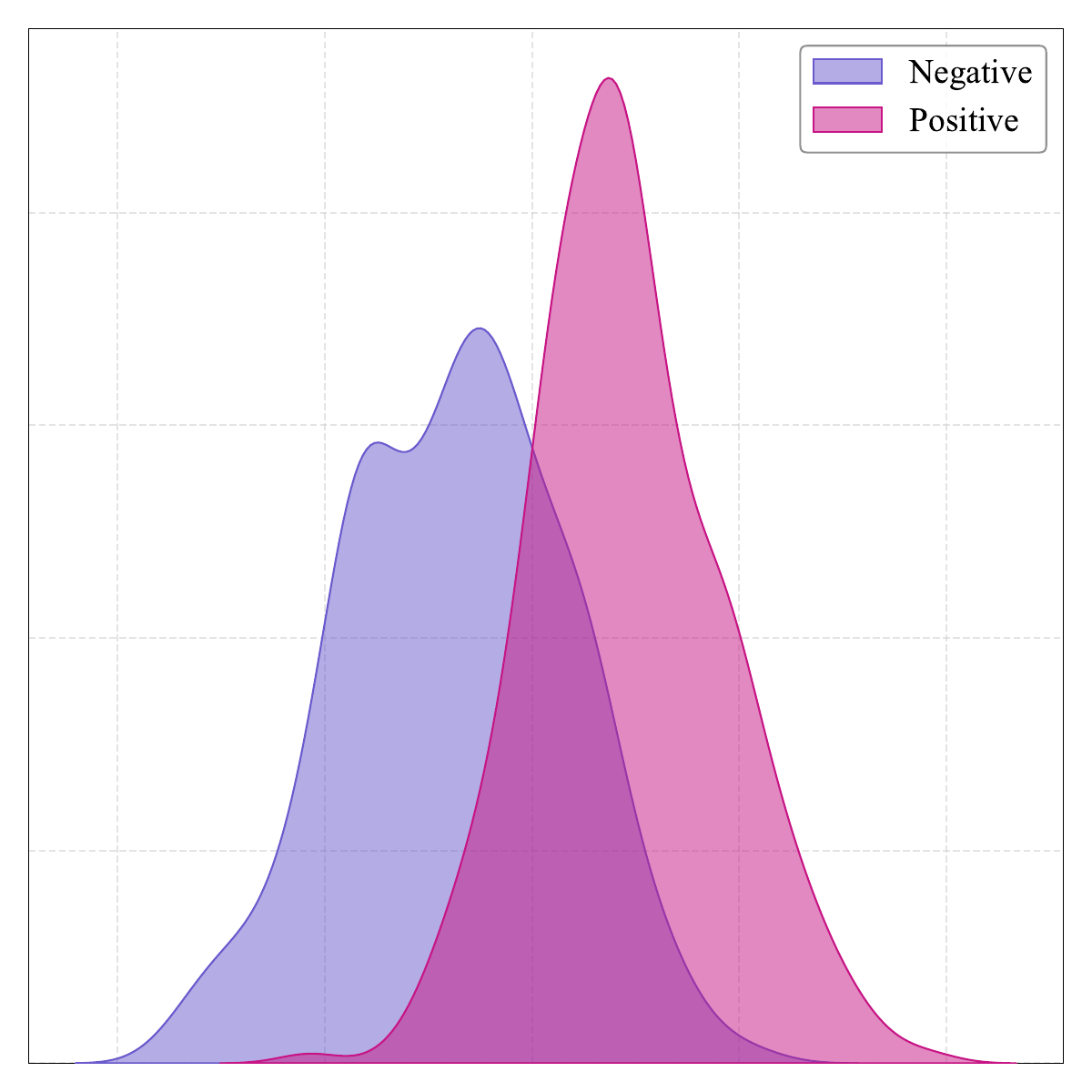}
						\end{subfigure} \\
									
						\rotatebox[origin=c]{90}{\small \textbf{Ambient Stream}} & 
						\begin{subfigure}[c]{0.44\columnwidth}
							\includegraphics[width=\linewidth]{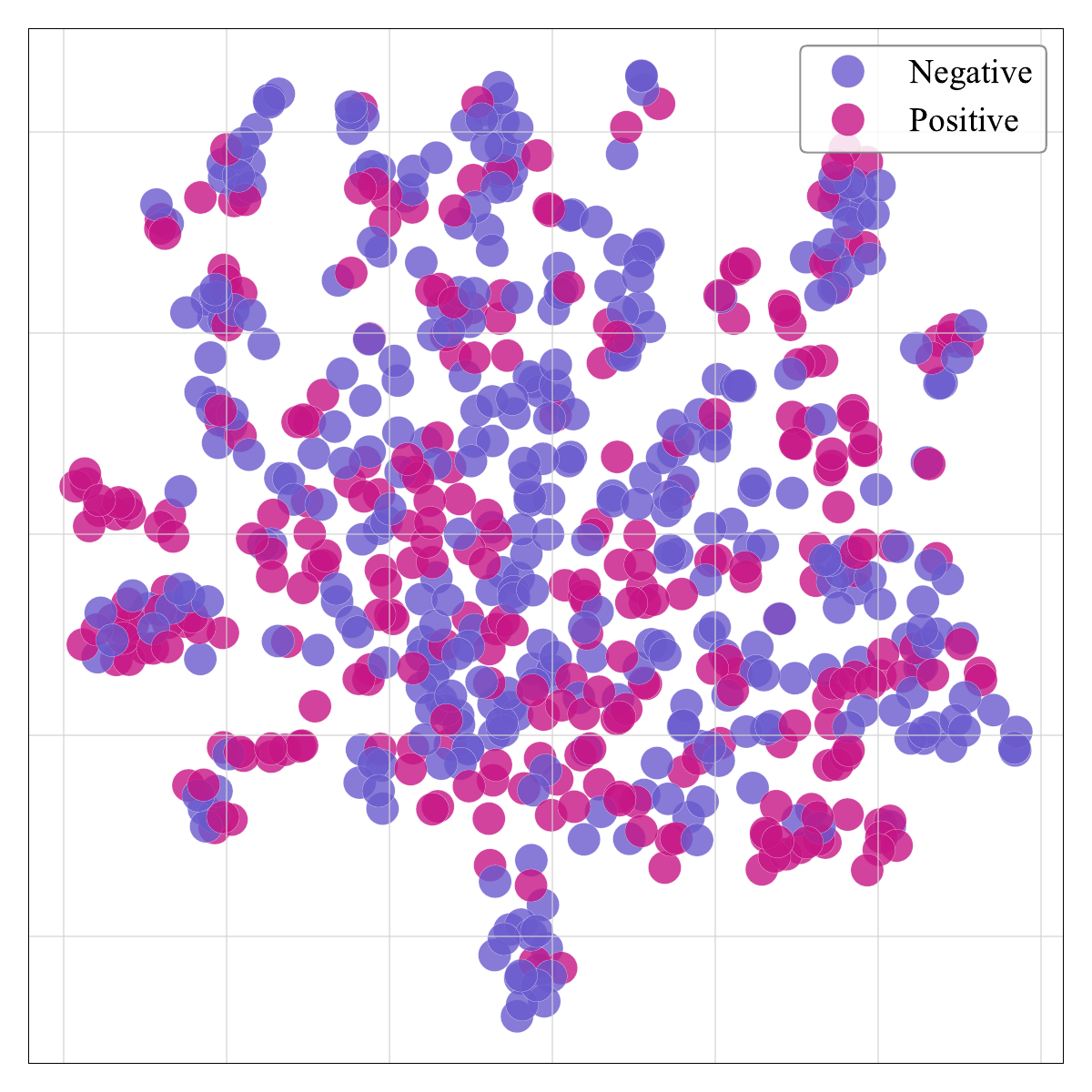}
						\end{subfigure} & 
						\begin{subfigure}[c]{0.44\columnwidth}
							\includegraphics[width=\linewidth]{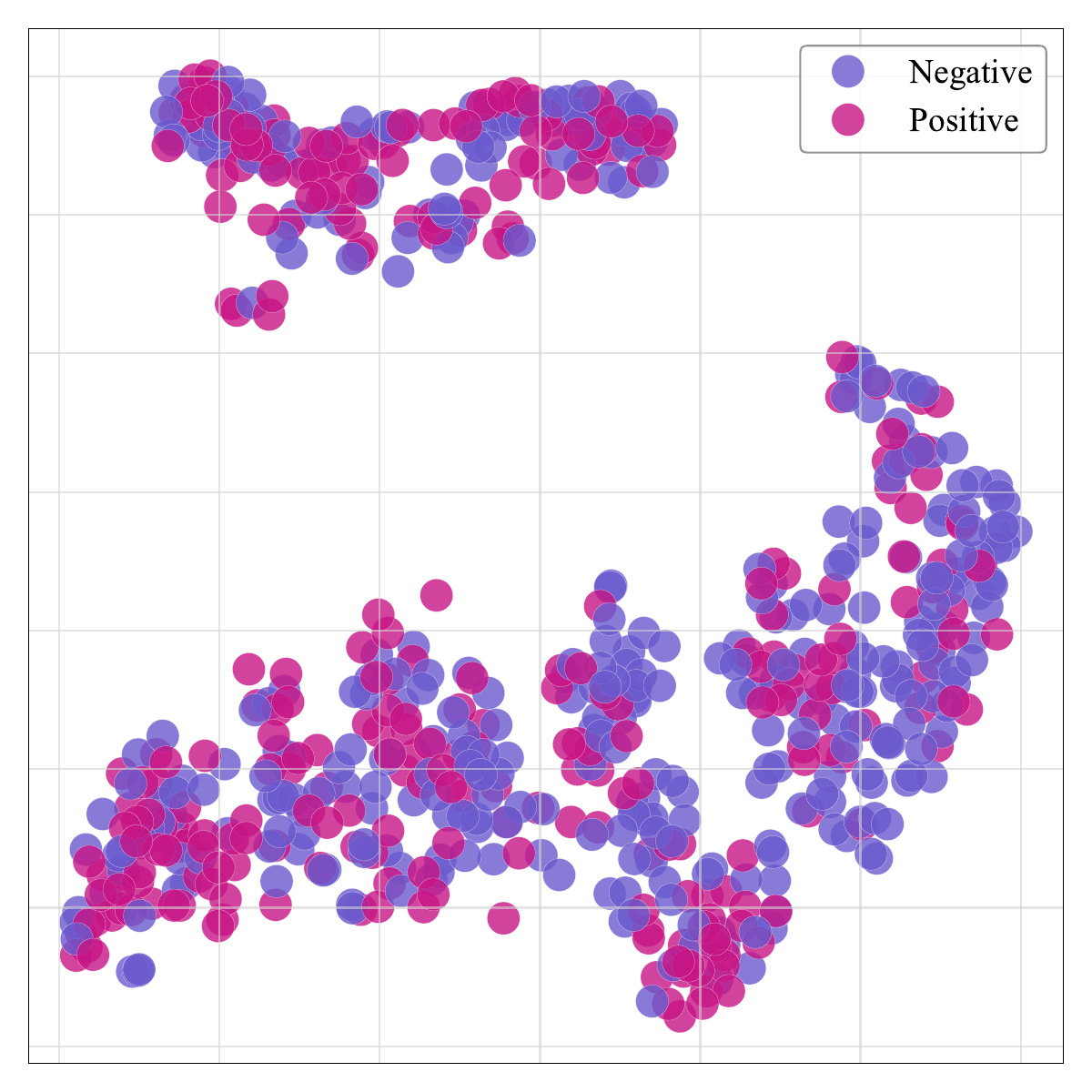}
						\end{subfigure}
					\end{tabular}
				}
				\caption{Visualization of feature disentanglement on the MOSI dataset.}
				\label{fig:disentanglement_analysis}
			\end{figure}
			
			\noindent\textbf{Visualization of Learned Representation Space.} In Figure~\ref{fig:tsne_visualization_8grid}, we visualize the t-SNE distribution \cite{linderman2017efficient} of the final [CLS] token on all four datasets. The baseline consistently exhibits a "chaotic cloud" distribution with severe class entanglement, indicating weak semantic discriminability. In strong contrast, our method reconstructs the feature space into a highly structured manifold across both languages. Specifically, on MOSI and MOSEI, we observe a clear continuous gradient transitioning from highly negative (red) to highly positive (purple), reflecting the model's precise grasp of fine-grained sentiment intensity. Similarly, on the Chinese CH-SIMS and CH-SIMS v2, the sentiment clusters become significantly more compact and distinguishable. This distinct clustering confirms that our method successfully maps raw information into a semantically separable space, reducing classifier ambiguity.

			\noindent\textbf{Visualization of Dual-Stream Feature Distribution.} In Figure~\ref{fig:disentanglement_analysis}, we qualitatively validate the dual-stream mechanism on the MOSI dataset. The focus stream exhibits strong linear separability under LDA projection, confirming the effective disentanglement of sentiment semantics. Conversely, the ambient stream shows a sentiment-agnostic distribution under t-SNE projection, effectively absorbing background states without leaking sentiment information.
			
\begin{figure}[tp]
				\centering
				\resizebox{0.95\columnwidth}{!}{%
					\begin{minipage}{\textwidth}
						\centering
						\begin{subfigure}[b]{0.48\columnwidth}
							\includegraphics[width=\linewidth]{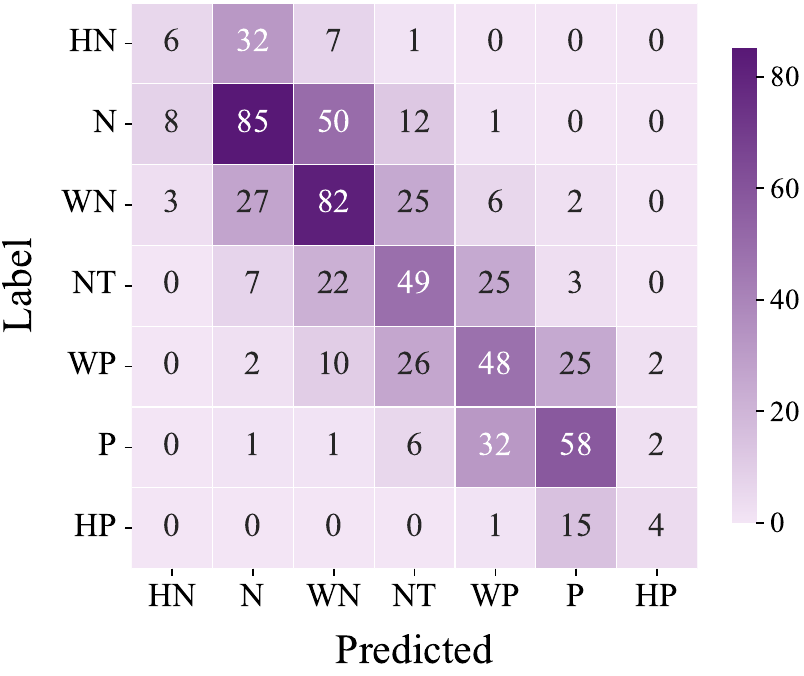}
						\end{subfigure}
						\hfill
						\begin{subfigure}[b]{0.48\columnwidth}
							\includegraphics[width=\linewidth]{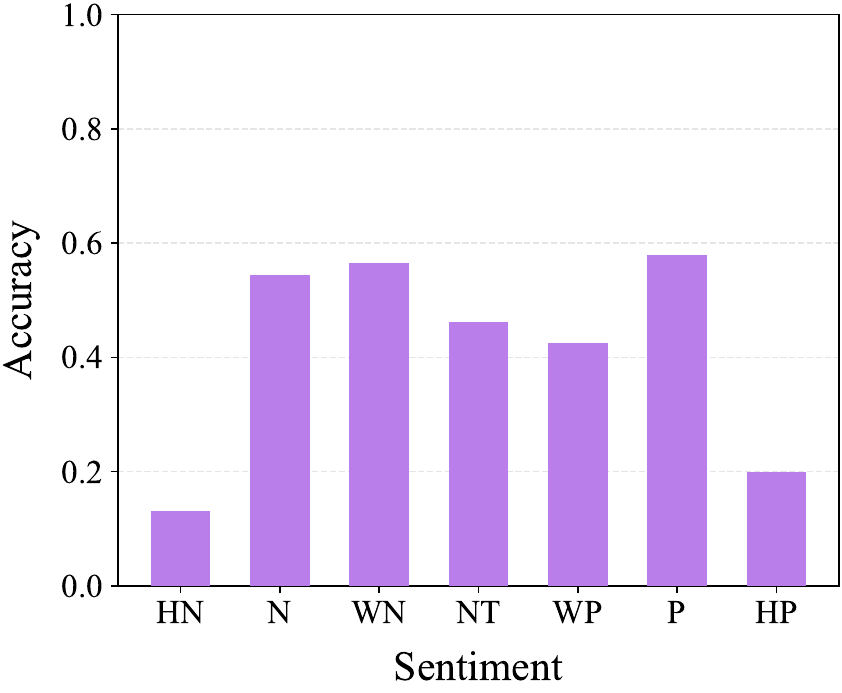}
						\end{subfigure}
					\end{minipage}%
				}
				\caption{
					Confusion matrix and per-class accuracy for 7-class sentiment prediction on the MOSI dataset
				}
				\label{fig:confusion_7class}
			\end{figure}

			\noindent\textbf{Analysis of Fine-grained Sentiment Prediction.} As shown in Figure~\ref{fig:confusion_7class}, we present the confusion matrix and corresponding accuracy of 7-class sentiment for the MOSI dataset. We observe that the samples for "HN" and "HP" yield the worst performance. The confusion matrix reveals these classes are relatively less than other sentiments, which indicates that the long-tail distribution limits performance on extreme sentiments.

	\section{Conclusion}\label{sec5}
	
		Grounded in the structural isomorphism between non-verbal modalities and natural language, we explore the potential of LLMs for multimodal sentiment analysis. However, the appearance-semantic misalignment hinders effective integration within the LLM semantic space. To address this, we propose SentiLLM, a framework incorporating a Dual-Stream Salience-Context Calibration Mechanism. This framework decouples audio-visual sequences into a focus stream containing salient sentiment shifts and an ambient stream representing contextual states. Extensive experiments across four datasets confirm the effectiveness of our proposed method. We anticipate this structural abstraction paradigm will offer novel insights into non-verbal modality sequence modeling.

\begin{acks}
	This work is supported by the National Natural Science Foundation of China (Grant No. 62272188). 
\end{acks}

\bibliographystyle{ACM-Reference-Format}
\balance
\bibliography{mfp5361-arxiv}

\clearpage
\appendix

\section*{Appendix}
	
\section{Experimental Details}\label{app:1}
	
	\subsection{Datasets}\label{app:11}
	
		In this section, we briefly introduce the datasets: 
		
		\begin{enumerate}
			\item \textbf{MOSI} \cite{Zadeh2016MOSIMC}: MOSI contains 2,199 segments from 93 videos. Speakers express personal opinions in a monologue format. Annotations cover a 7-point sentiment scale from -3 (strongly negative) to +3 (strongly positive).
			
			\item \textbf{MOSEI} \cite{zadeh2018multimodal}: MOSEI is an extended version of MOSI with larger scale. It contains 22,856 segments across 25 topics from YouTube.
			
			\item \textbf{CH-SIMS} \cite{yu2020ch}: CH-SIMS is a Chinese dataset with both unimodal and multimodal annotations. It includes 2,281 segments from movies and TV shows, annotated on a 5-point scale (-2 to +2).
			
			\item \textbf{CH-SIMS v2} \cite{liu2022make}: CH-SIMS v2 extends CH-SIMS with data from 11 scenarios, including interviews and talk shows, to simulate real-world diversity. This demands precise multimodal integration for correct prediction.
			
		\end{enumerate}
		
		\subsection{Baseline Methods}\label{app:12}
		
		We compare the proposed SentiLLM with the following competitive baselines:
		
		\begin{enumerate}
			
			\item \textbf{Tensor Fusion Network (TFN)} \cite{Zadeh2017TensorFN}: TFN explicitly captures unimodal, bimodal, and trimodal interactions via mathematical outer products. 
			
			\item \textbf{Low-rank Multimodal Fusion (LMF)} \cite{liu2018efficient}: LMF addresses TFN complexity by utilizing Low-rank Tensor Decomposition to reduce parameters from exponential to linear scale. 
			
			\item \textbf{Graph Memory Fusion Network (Graph-MFN)} \cite{zadeh2018multimodal}: Graph-MFN introduces a dynamic fusion graph to adaptively adjust edge weights between vertices, selecting effective fusion paths.
			
			\item \textbf{Multimodal Transformer (MuLT)} \cite{tsai2019multimodal}: MuLT abandons alignment and fusion strategy, utilizing cross-modal attention to directly capture long-range dependencies. 
			
			\item \textbf{MISA} \cite{Hazarika2020MISAMA}: MISA projects modalities into shared and specific sub-spaces, enforcing separation via orthogonality constraints and reconstruction losses.
			
			\item \textbf{Multimodal Adaptation Gate BERT (MAG-BERT)} \cite{rahman2020integrating}: MAG-BERT injects multimodal information into BERT via a soft adaptation gate, refining text representations.
			
			\item \textbf{Self-Supervised Multi-Task Learning (Self-MM)} \cite{yu2021learning}: Self-MM generates pseudo-labels from multimodal targets to construct self-supervised multi-task learning.
			
			\item \textbf{Multimodal Mutual Information Maximization (MMIM)} \cite{han2021improving}: MMIM maximizes mutual information between fusion results and unimodal inputs to preserve key modality-specific information.
			
			\item \textbf{Decoupled Multimodal Distillation (DMD)} \cite{li2023decoupled}: DMD decomposes representations into modality-agnostic and modality-specific parts via knowledge distillation.
			
			\item \textbf{MLP-Based Fusion (CubeMLP)} \cite{sun2022cubemlp}: CubeMLP treats data as a 3D cube, mixing features across temporal, modality, and channel axes using a pure MLP architecture.
			
			\item \textbf{Audio-Visual Mixup Consistency (AV-MC)} \cite{liu2022make}: AV-MC introduces Mixup strategies to mitigate model over-reliance on the text modality.
			
			\item \textbf{Disentangled-Language-Focused Framework (DLF)} \cite{wang2025dlf}: DLF revisits disentanglement with a language-dominant perspective, using language queries to aggregate specific information from other modalities.
			
			\item \textbf{KAN-MCP} \cite{luo2025towards}: KAN-MCP places activation functions on edges using KANs, making the fusion process transparent and interpretable.
			
			\item \textbf{Unified MSA and ERC (UniMSE)} \cite{hu2022unimse}: UniMSE uses T5 as a backbone, incorporating cross-modal contrastive learning guided by syntax dependency trees.
			
			\item \textbf{UniSA} \cite{li2023unisa}: UniSA converts audio-visual features into discrete tokens via clustering for LLM generation.
			
			\item \textbf{MSE-Adapter} \cite{yang2025mse}: MSE-Adapter builds a Text-Guide-Mixer to explicitly connect text tokens with non-text features for LLM analysis.
			
		\end{enumerate}
	
	\subsection{LLM Backbones}\label{app:13}
	
		We compare the proposed SentiLLM with the following competitive baselines:
		
		\begin{enumerate}
			\item \textbf{Qwen2.5-1.5B} \cite{Yang2024Qwen25TR}: Qwen2.5-1.5B achieves exceptional instruction following with minimal parameters, leveraging 18T tokens and Grouped-Query Attention.
			
			\item \textbf{Llama-3.2-3B} \cite{dubey2024llama}: Llama-3.2-3B is distilled and pruned from a 70B Llama-3 model, inheriting complex logical reasoning capabilities.
			
			\item \textbf{Gemma-2-2B-it} \cite{team2024gemma}: Gemma-2-2B-it combines sliding windows and global attention to capture both local details and global context.
			
			\item \textbf{InternLM2.5-1.8B} \cite{cai2024internlm2}: InternLM2.5-1.8B features strong long-context processing, making it ideal for analyzing temporal evolution in multi-turn dialogue.
		\end{enumerate}
		
	\subsection{Training Hyperparameters}\label{app:14}
	
		We adjusted hyperparameters based on dataset scale and complexity. Details are provided in Table~\ref{tab:hyperparams}.	
	
\section{Extended Ablation Studies}
	
	\subsection{Impact of Different Modalities}\label{app:modality_ablation}
	
		Table~\ref{tab:modality_ablation} validates the necessity of multimodal integration. The "text only" baseline performs decently due to LLM capabilities, but incorporating visual or audio signals brings significant gains. This indicates that non-verbal modalities carry critical sentiment information ignored by text, and our method effectively extracts text-aligned sentiment elements to activate LLM reasoning.
		
		\begin{table}[!t]
			\centering
			\caption{Hyperparameter settings across different datasets.}
			\label{tab:hyperparams}
			\renewcommand{\arraystretch}{1.2}
			\resizebox{\columnwidth}{!}{%
				\begin{tabular}{lcccc}
					\toprule
					\textbf{Parameter} & \textbf{MOSI} & \textbf{MOSEI} & \textbf{CH-SIMS} & \textbf{CH-SIMS v2} \\
					\midrule
					Optimizer             & \multicolumn{4}{c}{AdamW} \\
					Scheduler             & \multicolumn{4}{c}{Cosine Decay} \\
					\midrule
					Learning Rate         & 1e-3 & 2e-3 & 5e-4 & 1e-3 \\
					Weight Decay          & 0.01 & 0.01 & 0.05 & 0.05 \\
					Warmup Ratio          & 0.10 & 0.20 & 0.10 & 0.10 \\
					\midrule
					Batch Size            & 16   & 16   & 16   & 16   \\
					Gradient Accum       & 2    & 2    & 2    & 2    \\
					Total Epochs          & 50   & 30   & 40   & 30   \\
					\midrule
					Sparsity Ratio $k$    & 0.6  & 0.4  & 0.6  & 0.6  \\
					\bottomrule
				\end{tabular}
			}
		\end{table}
		
		\begin{table}[!t]
			\centering
			\caption{Ablation study on modality impact on the MOSI dataset.}
			\label{tab:modality_ablation}
			\resizebox{\columnwidth}{!}{%
				\begin{tabular}{l cccccc}
					\toprule
					\textbf{Modalities} & \textbf{MAE} ($\downarrow$) & \textbf{Corr} ($\uparrow$) & \textbf{Acc-7} ($\uparrow$) & \textbf{Acc-5} ($\uparrow$) & \textbf{Acc-2} ($\uparrow$) & \textbf{F1} ($\uparrow$) \\
					\midrule
					Text Only & 0.760 & 0.793 & 39.50 & 47.67 & 84.30 & 84.21 \\
					Text + Video & 0.676 & 0.831 & 43.59 & 52.19 & 88.11 & 88.08 \\
					Text + Audio & {0.648} & {0.841} & {46.65} & {54.08} & {88.57} & {88.53} \\
					\midrule
					\textbf{Text + Video + Audio} & \textbf{0.640} & \textbf{0.846} & \textbf{48.40} & \textbf{56.71} & \textbf{89.02} & \textbf{88.99} \\
					\bottomrule
				\end{tabular}%
			}
		\end{table}
		
		\begin{table}[!t]
			\centering
			\caption{Comparative analysis of Soft Salience Disentanglement and Hard Truncation. Sparsity ratio $k$ is set to 0.6.}
			\label{tab:soft_vs_hard}
			\resizebox{\columnwidth}{!}{%
				\begin{tabular}{l cccccc}
					\toprule
					\textbf{Gating Strategy} & \textbf{MAE} ($\downarrow$) & \textbf{Corr} ($\uparrow$) & \textbf{Acc-7} ($\uparrow$) & \textbf{Acc-5} ($\uparrow$) & \textbf{Acc-2} ($\uparrow$) & \textbf{F1} ($\uparrow$) \\
					\midrule
					Hard Truncation & {0.664} & {0.835} & {46.79} & {54.23} & {87.96} & {87.95} \\
					\midrule
					\textbf{Soft Salience} & \textbf{0.640} & \textbf{0.846} & \textbf{48.40} & \textbf{56.71} & \textbf{89.02} & \textbf{88.99} \\
					\bottomrule
				\end{tabular}%
			}
		\end{table}
		
		\begin{table}[!t]
			\centering
			\caption{Ablation study on the effectiveness of sparsity loss on the MOSI dataset.}
			\label{tab:sparsity_loss}
			\resizebox{\columnwidth}{!}{%
				\begin{tabular}{l cccccc}
					\toprule
					\textbf{Loss Function} & \textbf{MAE} ($\downarrow$) & \textbf{Corr} ($\uparrow$) & \textbf{Acc-7} ($\uparrow$) & \textbf{Acc-5} ($\uparrow$) & \textbf{Acc-2} ($\uparrow$) & \textbf{F1} ($\uparrow$) \\
					\midrule
					w/o Sparsity Loss & {0.691} & {0.830} & {42.57} & {51.17} & {88.41} & {88.38} \\
					\midrule
					\textbf{w/ Sparsity Loss} & \textbf{0.640} & \textbf{0.846} & \textbf{48.40} & \textbf{56.71} & \textbf{89.02} & \textbf{88.99} \\
					\bottomrule
				\end{tabular}%
			}
		\end{table}	
		
		\begin{table}[!t]
			\centering
			\caption{Ablation study on computational efficiency and overhead for single-batch inference}
			\label{tab:model_efficiency}
			\footnotesize
			\setlength{\tabcolsep}{4.5pt}
			\begin{tabular}{l ccccc}
				\toprule
				\textbf{Setting} & 
				\makecell[c]{\textbf{FLOPs} \\ \textbf{(G)}} & 
				\makecell[c]{\textbf{Memory} \\ \textbf{(MB)}} & 
				\makecell[c]{\textbf{Latency} \\ \textbf{(ms)}} & 
				\makecell[c]{\textbf{Throughput} \\ \textbf{(samples/s)}} & 
				\makecell[c]{\textbf{Params} \\ \textbf{(M)}} \\
				\midrule
				Vanilla SentiLLM	& 2725.21 & 6170.36 & 44.48 & 359.68 & 1543.91 \\
				\textbf{SentiLLM (Ours)}        & 3050.76 & 6241.85 & 53.43 & 299.46 & 1548.59 \\
				\midrule[\cmidrulewidth] 
				Incremental ($\Delta$) & +325.55 &  +71.49 & +8.95 & --     &   +4.68 \\
				\bottomrule
			\end{tabular}
		\end{table}

	\subsection{Soft Salience Disentanglement and Hard Truncation}\label{app:gating_mechanism}
	
		As shown in Table~\ref{tab:soft_vs_hard}, our soft salience disentanglement outperforms hard truncation across all metrics. This indicates that soft salience disentanglement maintains differentiability and assigns continuous salience scores, ensuring flexible separation between focus and ambient streams.
	
	\subsection{Effectiveness of Sparsity Loss}\label{app:sparsity_loss}
	
		Table~\ref{tab:sparsity_loss} highlights the critical role of sparsity loss. A sharp drop in fine-grained accuracy is observed when removing this constraint. This suggests that salience scores converge to trivial solutions without explicit regularization. And the sparsity loss forces the model to make decisive binary distinctions, ensuring meaningful separation between the two streams.
		
	\subsection{Model Efficiency}\label{app:24}
		
		To substantiate our parameter-efficient claim, single-batch inference tests on an RTX 5090 (Table~\ref{tab:model_efficiency}) demonstrate that SentiLLM introduces minimal overhead. This confirms its viability as a lightweight reasoning enhancer.
	
\section{Detailed Experiment Results}\label{app:detail_exp}

	\subsection{Semantic Sentences}\label{app:15}

		As shown in Table~\ref{tab:tasq_prompts_mosi} and ~\ref{tab:tasq_prompts_sims}, we detail the natural language prompts used in our semantic sentence setting. These sentence-level prompts cover multiple sentiment dimensions, including valence, intensity, and social cues, instead of using simplified keywords (positive/negative) or random vectors.	
	
	\subsection{Visualization of Dual-Stream Feature Distribution}\label{app:sne_ablation}
	
\begin{figure*}[b]
			\centering
			\begin{minipage}{0pt} \quad \end{minipage}
			\hfill
			\begin{minipage}{0.22\linewidth}\centering \small \textbf{MOSI} \end{minipage}\hfill
			\begin{minipage}{0.22\linewidth}\centering \small \textbf{MOSEI} \end{minipage}\hfill
			\begin{minipage}{0.22\linewidth}\centering \small \textbf{CH-SIMS} \end{minipage}\hfill
			\begin{minipage}{0.22\linewidth}\centering \small \textbf{CH-SIMS v2} \end{minipage}
			
			\vspace{2pt} 
			
			\rotatebox{90}{\makebox[0.50\columnwidth][c]{\small \textbf{Video (Focus Stream)}}}%
			\hspace{0pt}%
			\begin{subfigure}[b]{0.24\linewidth}
				\includegraphics[width=\textwidth]{./fig/scheme2_mosi_ours_Video_LDA.pdf}
			\end{subfigure}\hfill
			\begin{subfigure}[b]{0.24\linewidth}
				\includegraphics[width=\textwidth]{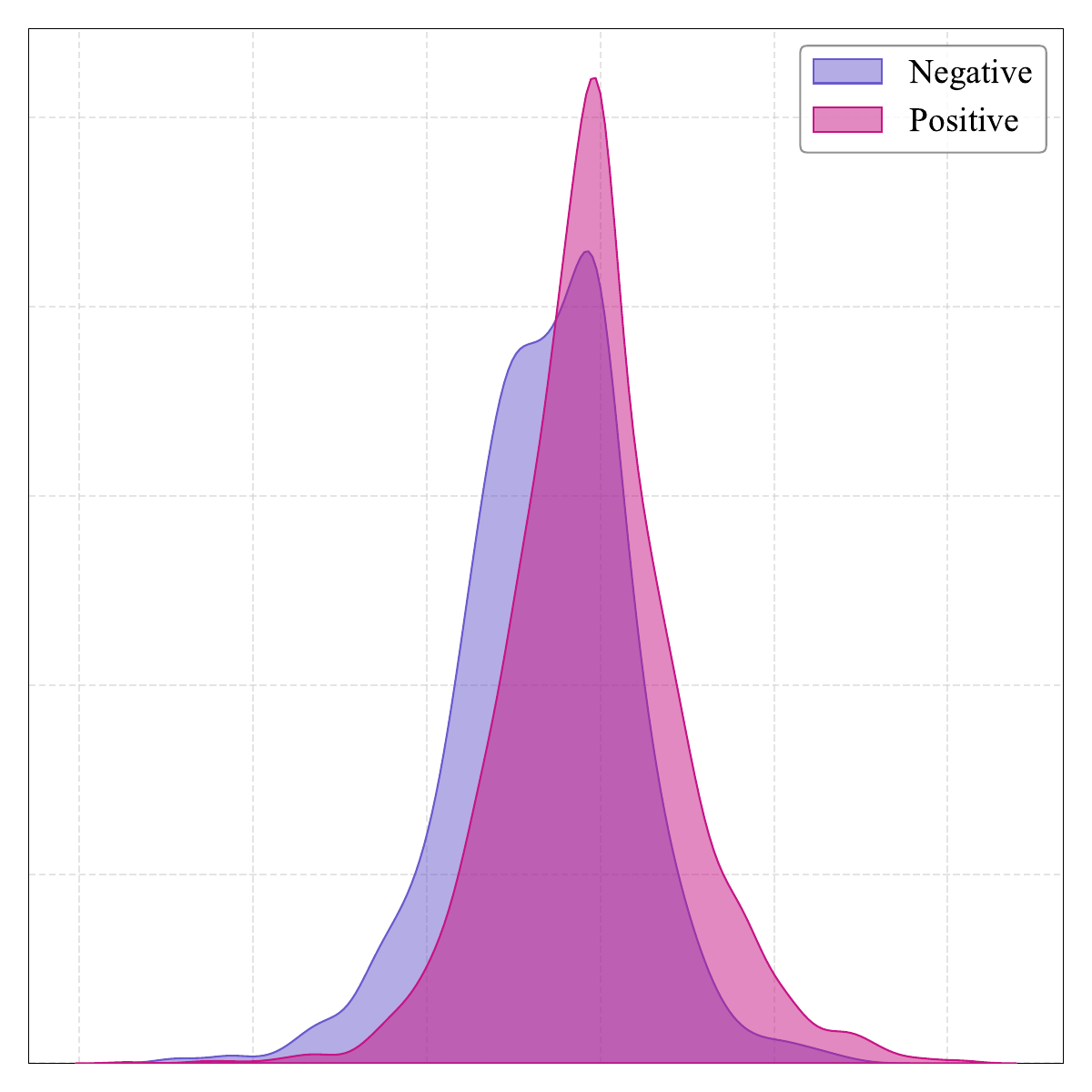}
			\end{subfigure}\hfill
			\begin{subfigure}[b]{0.24\linewidth}
				\includegraphics[width=\textwidth]{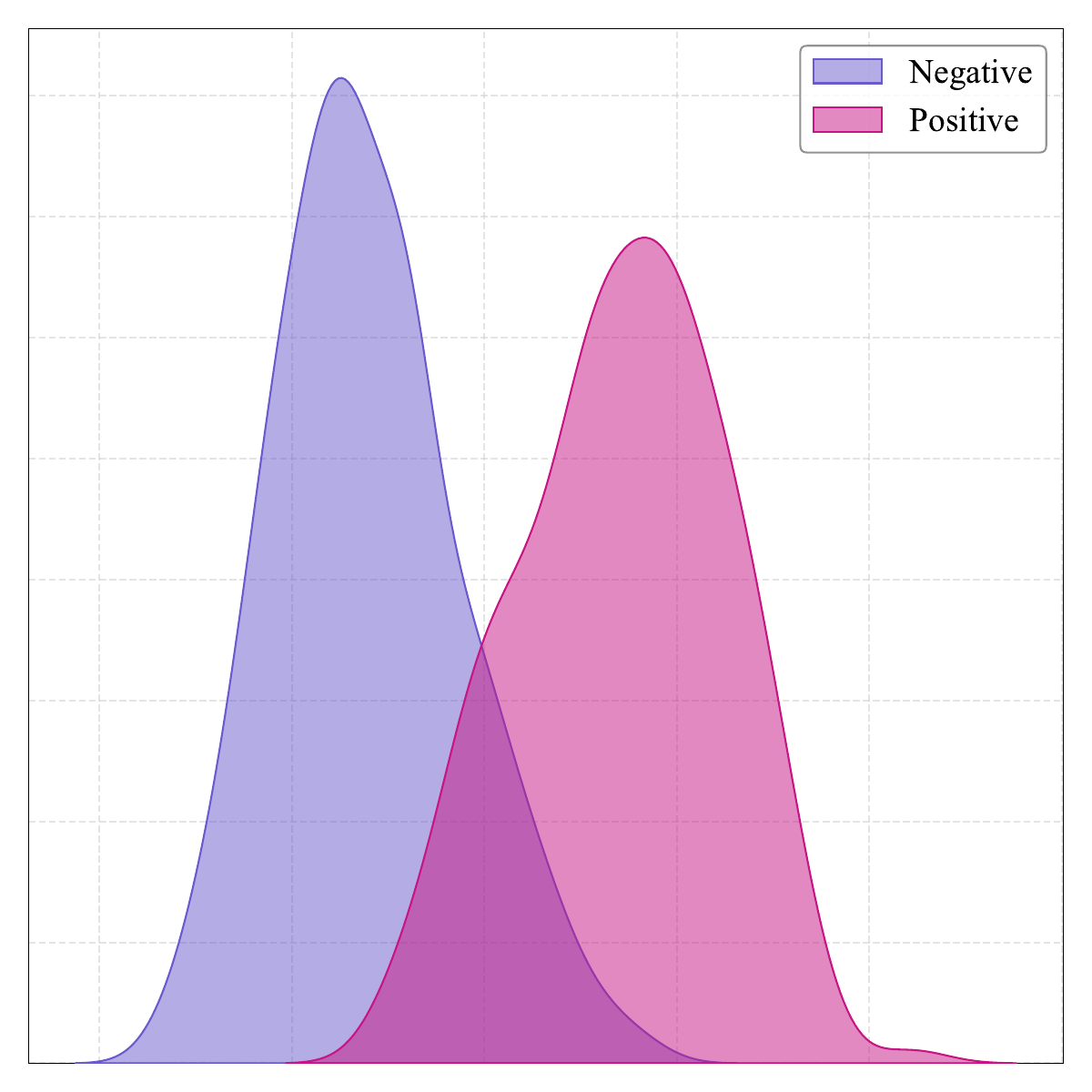}
			\end{subfigure}\hfill
			\begin{subfigure}[b]{0.24\linewidth}
				\includegraphics[width=\textwidth]{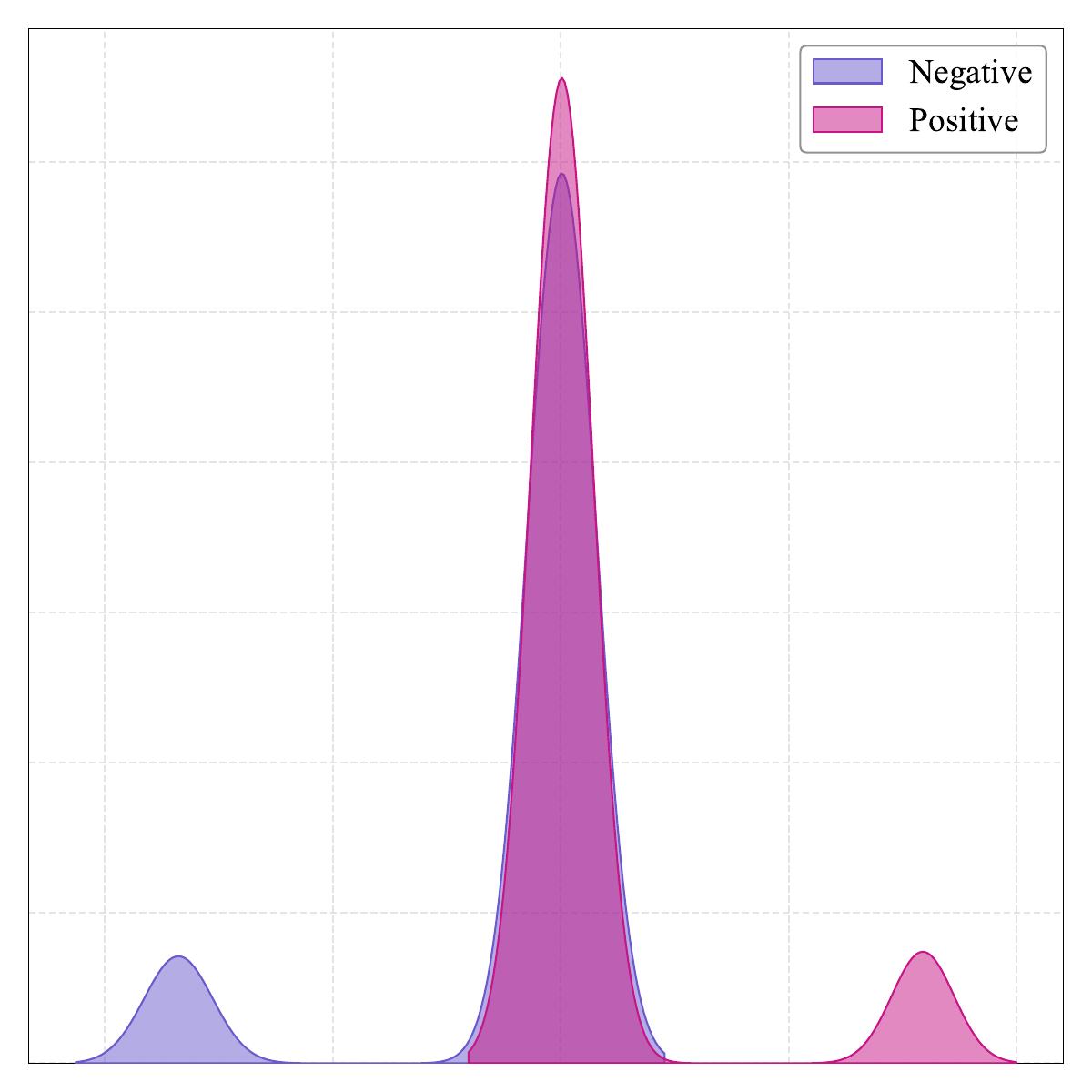}
			\end{subfigure}
			
			\vspace{2pt} 
			
			\rotatebox{90}{\makebox[0.50\columnwidth][c]{\small \textbf{Video (Ambient Stream)}}}%
			\hspace{0pt}
			\begin{subfigure}[b]{0.24\linewidth}
				\includegraphics[width=\textwidth]{./fig/scheme2_mosi_ours_Video_Ambient.pdf}
			\end{subfigure}\hfill
			\begin{subfigure}[b]{0.24\linewidth}
				\includegraphics[width=\textwidth]{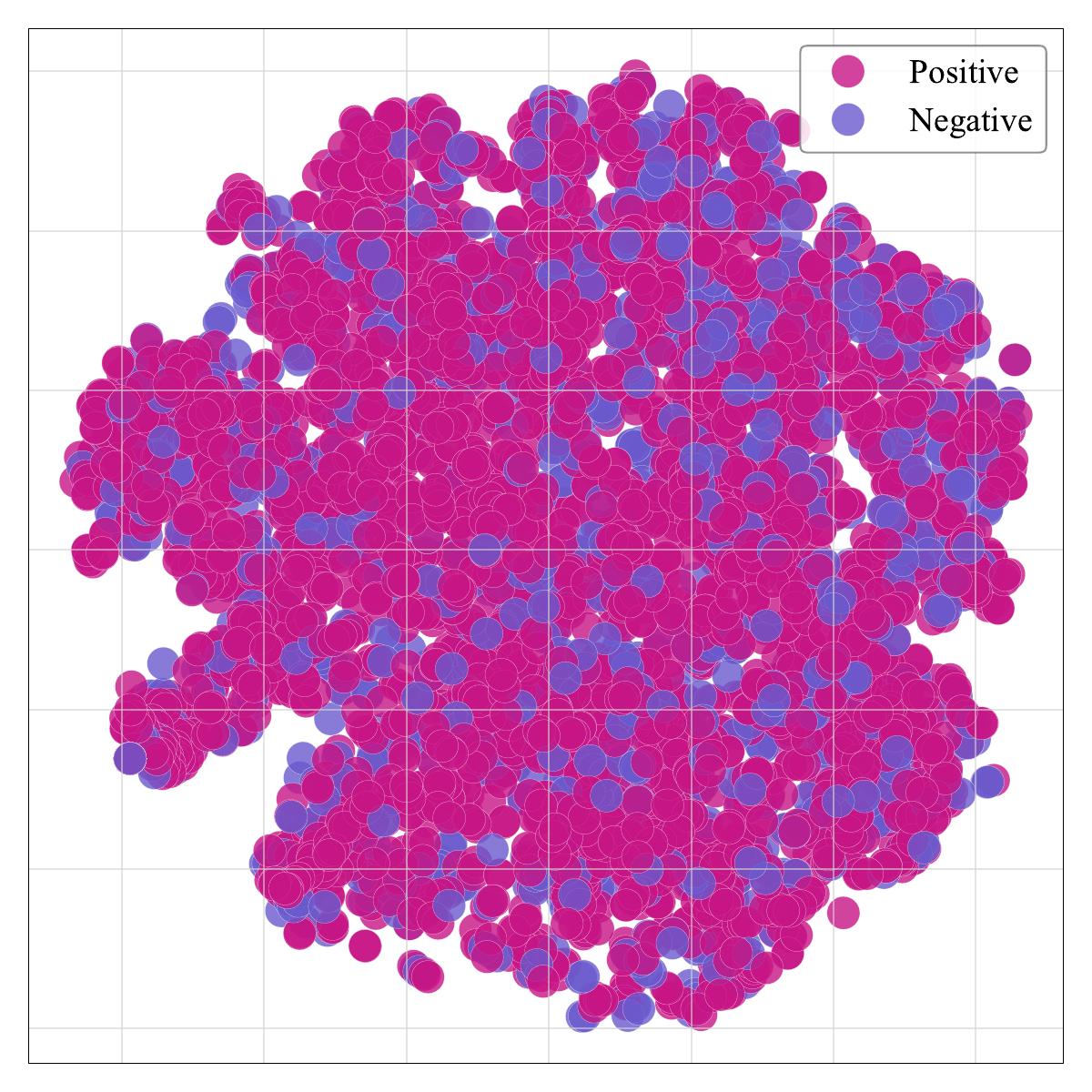}
			\end{subfigure}\hfill
			\begin{subfigure}[b]{0.24\linewidth}
				\includegraphics[width=\textwidth]{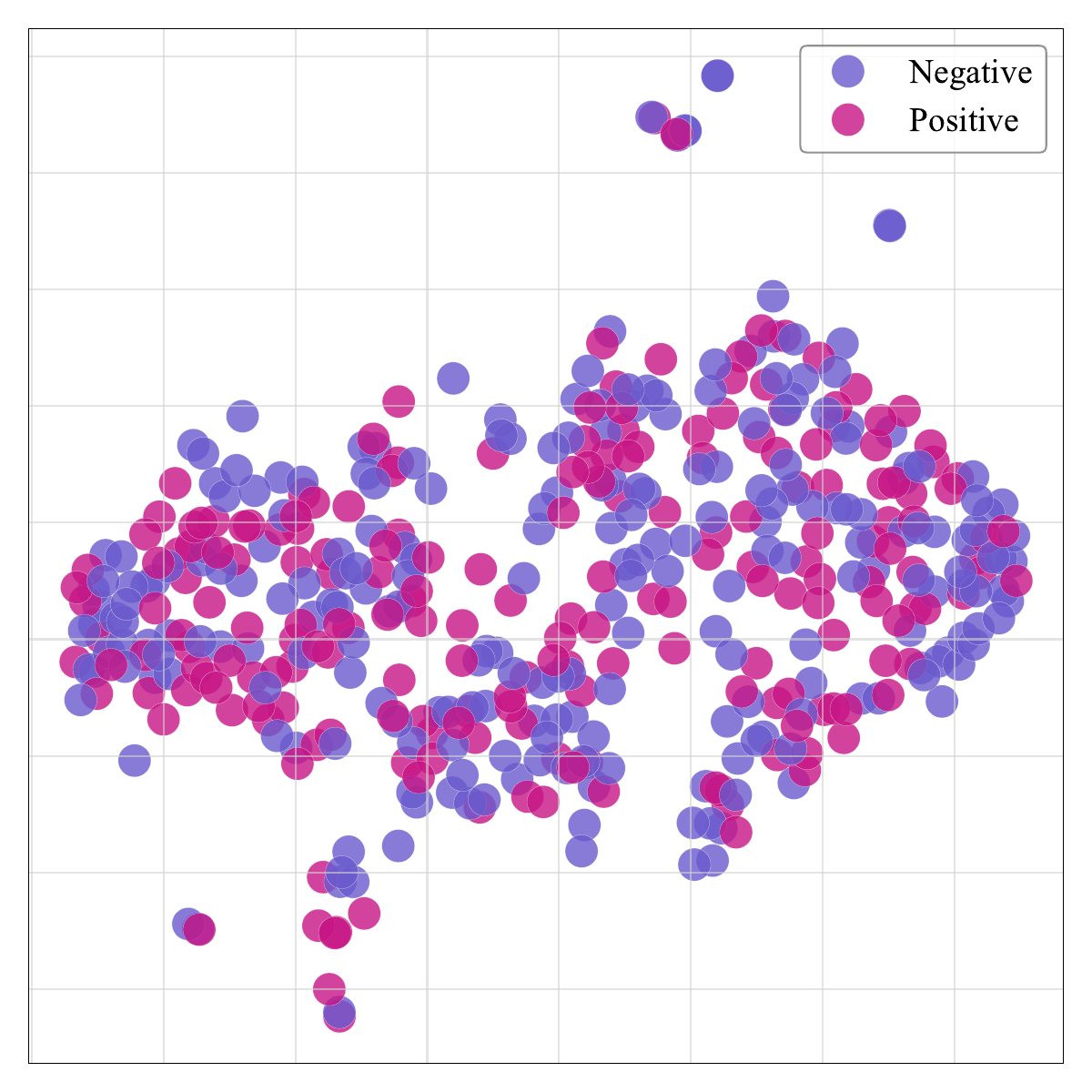}
			\end{subfigure}\hfill
			\begin{subfigure}[b]{0.24\linewidth}
				\includegraphics[width=\textwidth]{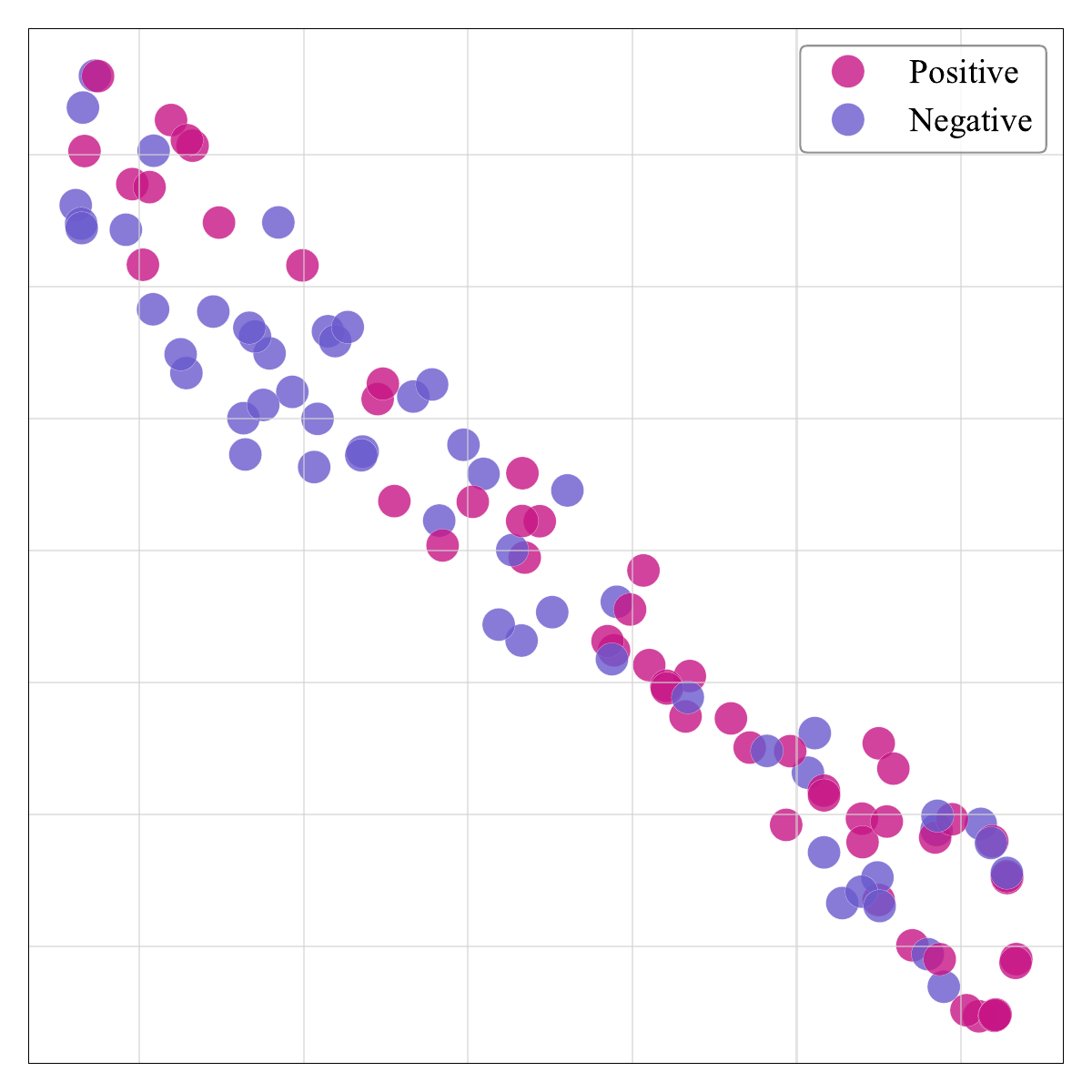}
			\end{subfigure}
			
			\vspace{2pt}
			
			\rotatebox{90}{\makebox[0.50\columnwidth][c]{\small \textbf{Audio (Focus Stream)}}}%
			\hspace{0pt}
			\begin{subfigure}[b]{0.24\linewidth}
				\includegraphics[width=\textwidth]{./fig/scheme2_mosi_ours_Audio_LDA.pdf}
			\end{subfigure}\hfill
			\begin{subfigure}[b]{0.24\linewidth}
				\includegraphics[width=\textwidth]{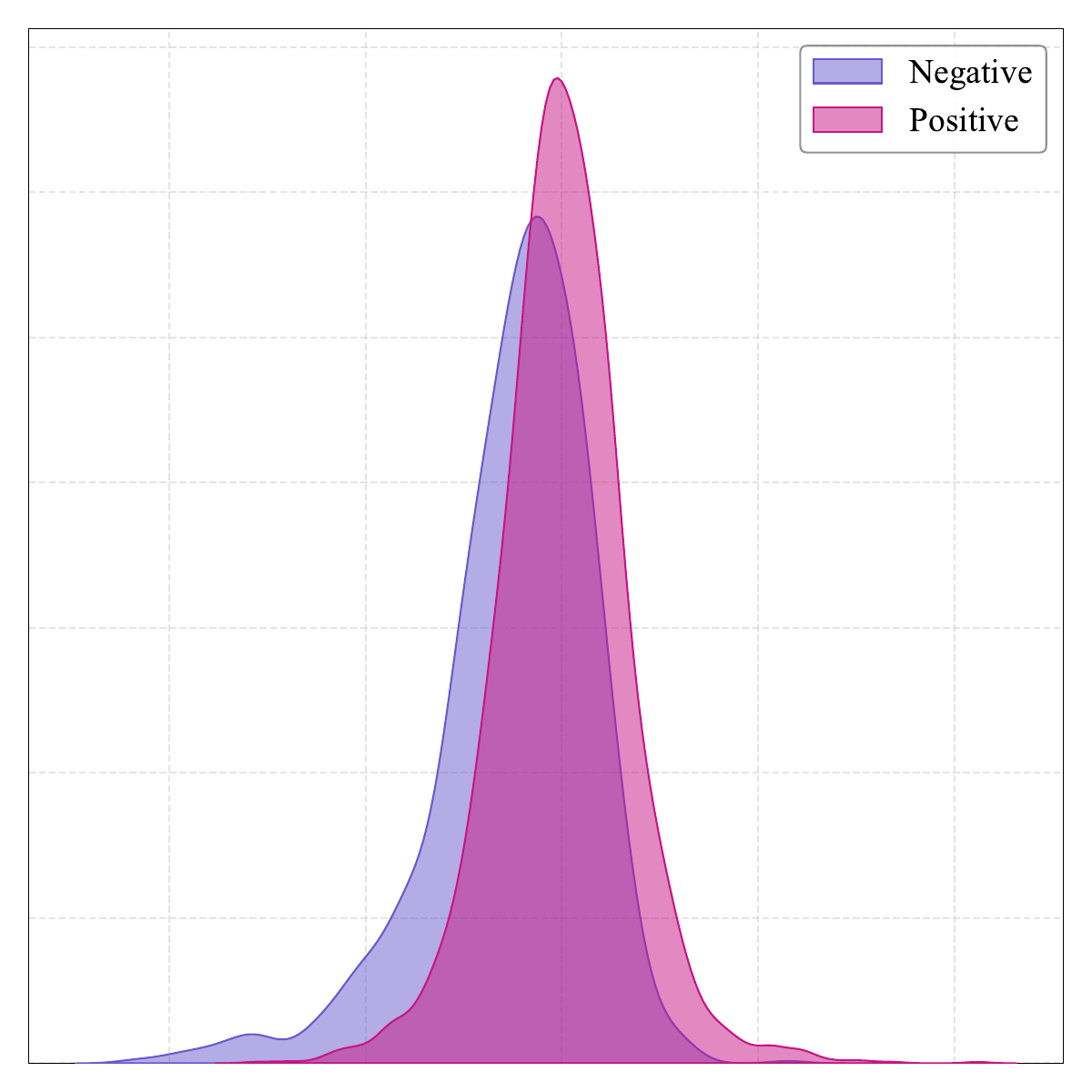}
			\end{subfigure}\hfill
			\begin{subfigure}[b]{0.24\linewidth}
				\includegraphics[width=\textwidth]{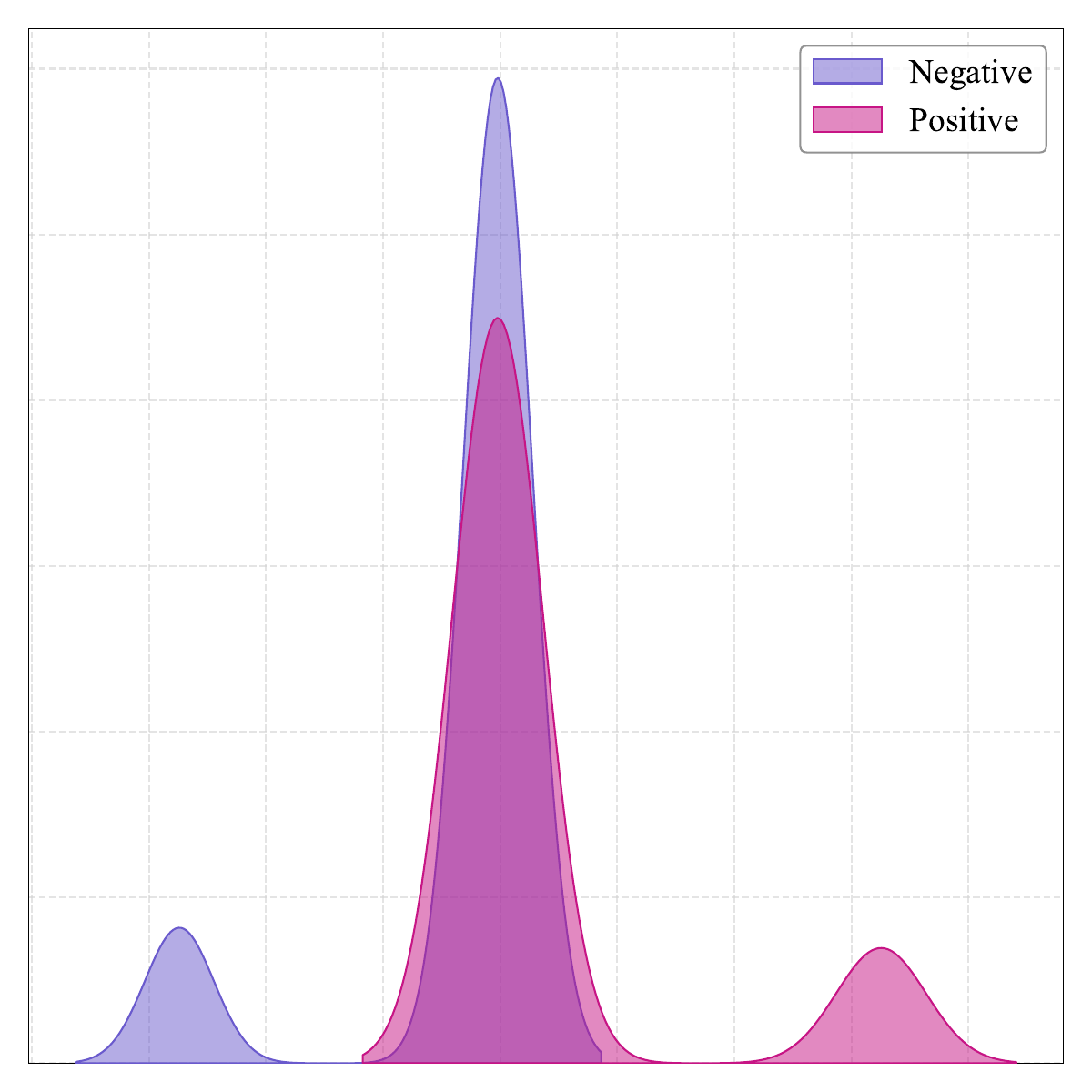}
			\end{subfigure}\hfill
			\begin{subfigure}[b]{0.24\linewidth}
				\includegraphics[width=\textwidth]{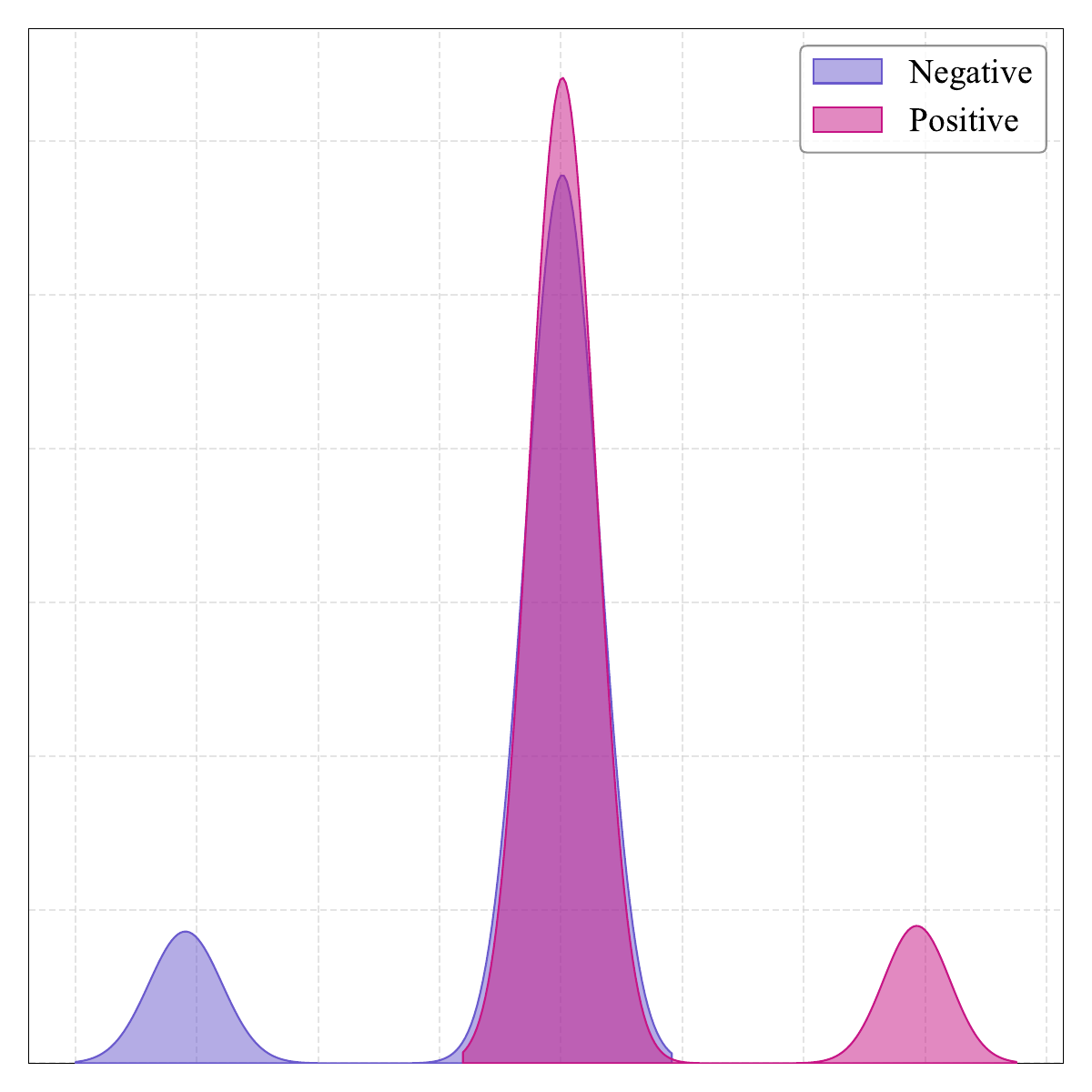}
			\end{subfigure}
			
			\vspace{2pt} 
			
			\rotatebox{90}{\makebox[0.50\columnwidth][c]{\small \textbf{Audio (Ambient Stream)}}}%
			\hspace{0pt}
			\begin{subfigure}[b]{0.24\linewidth}
				\includegraphics[width=\textwidth]{./fig/scheme2_mosi_ours_Audio_Ambient.pdf}
			\end{subfigure}\hfill
			\begin{subfigure}[b]{0.24\linewidth}
				\includegraphics[width=\textwidth]{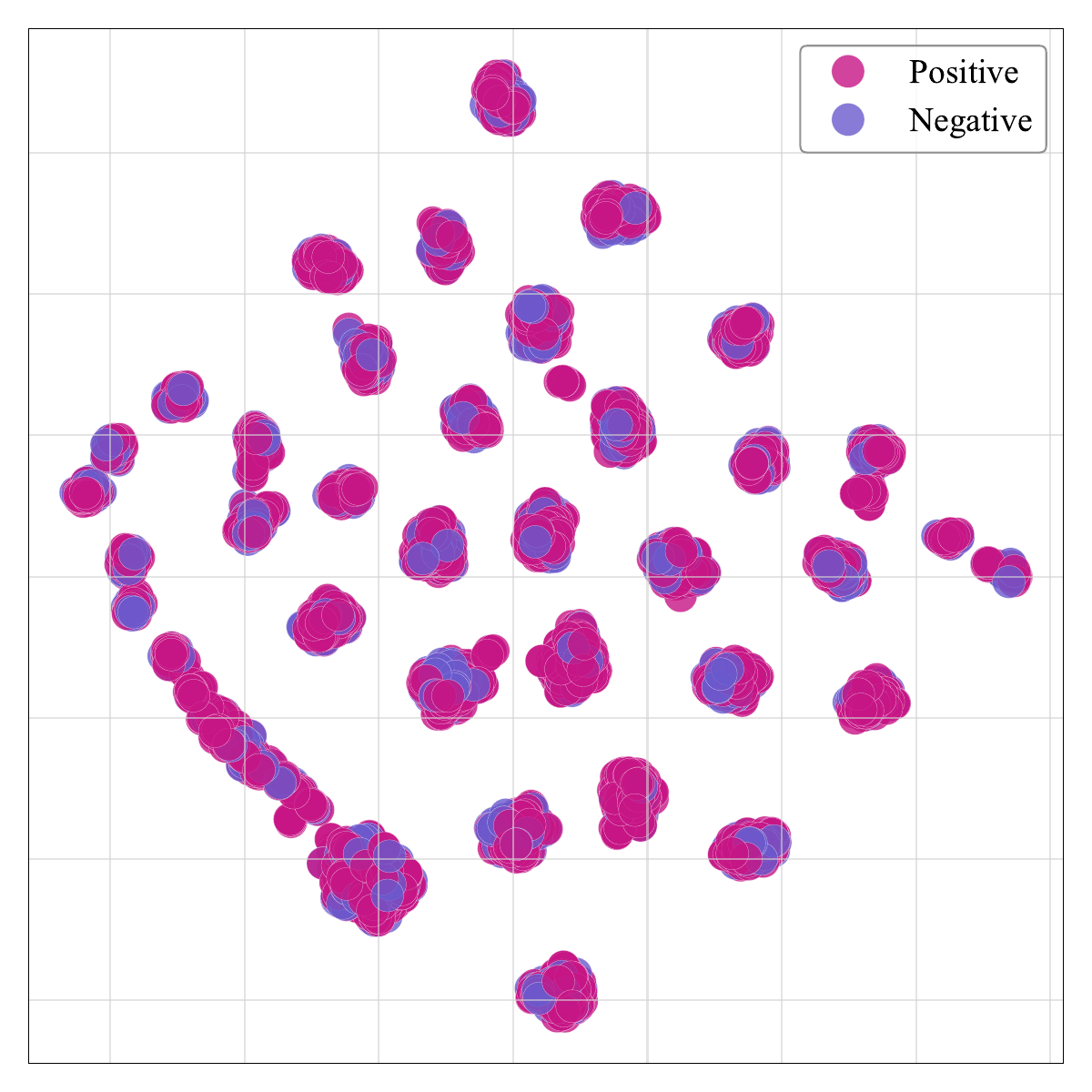}
			\end{subfigure}\hfill
			\begin{subfigure}[b]{0.24\linewidth}
				\includegraphics[width=\textwidth]{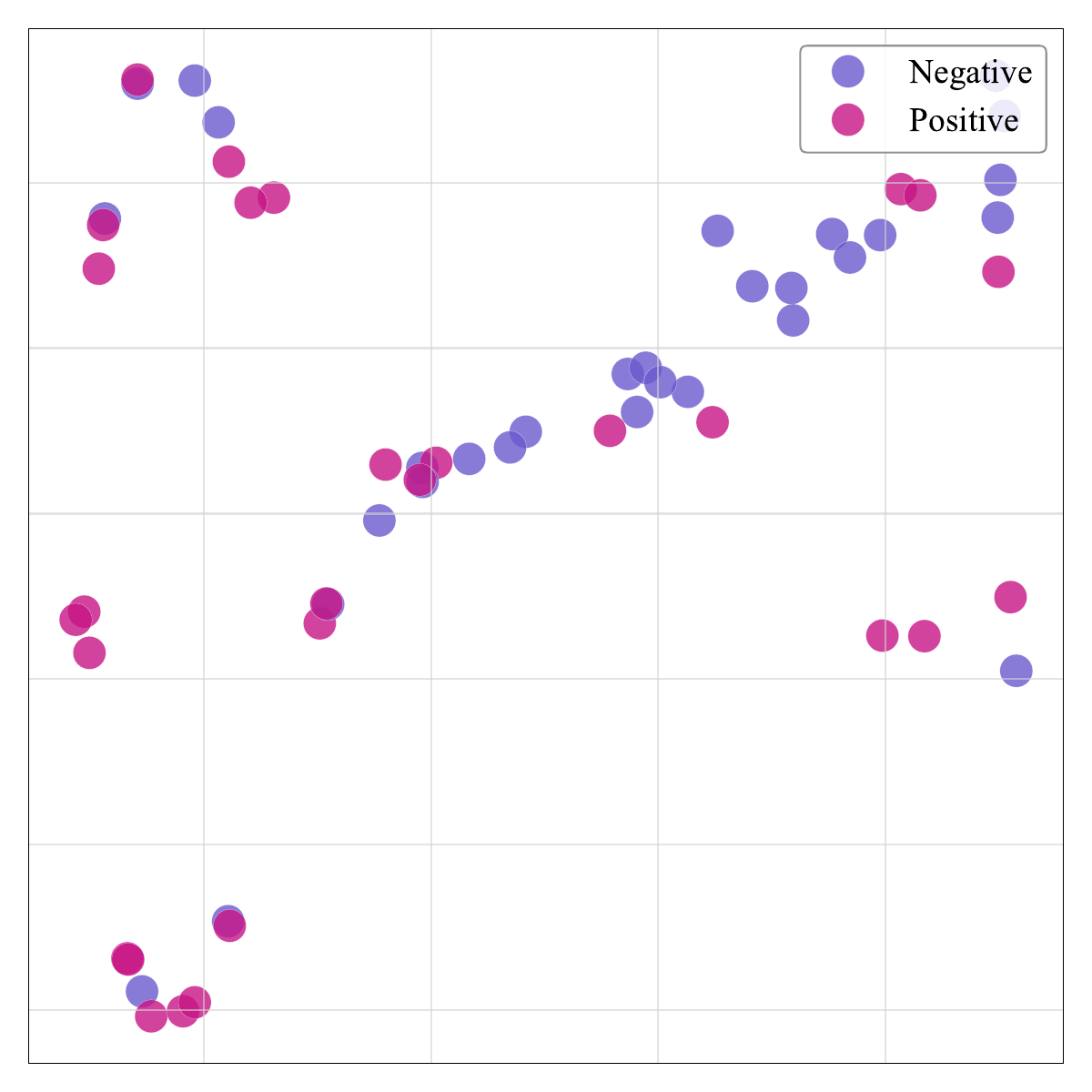}
			\end{subfigure}\hfill
			\begin{subfigure}[b]{0.24\linewidth}
				\includegraphics[width=\textwidth]{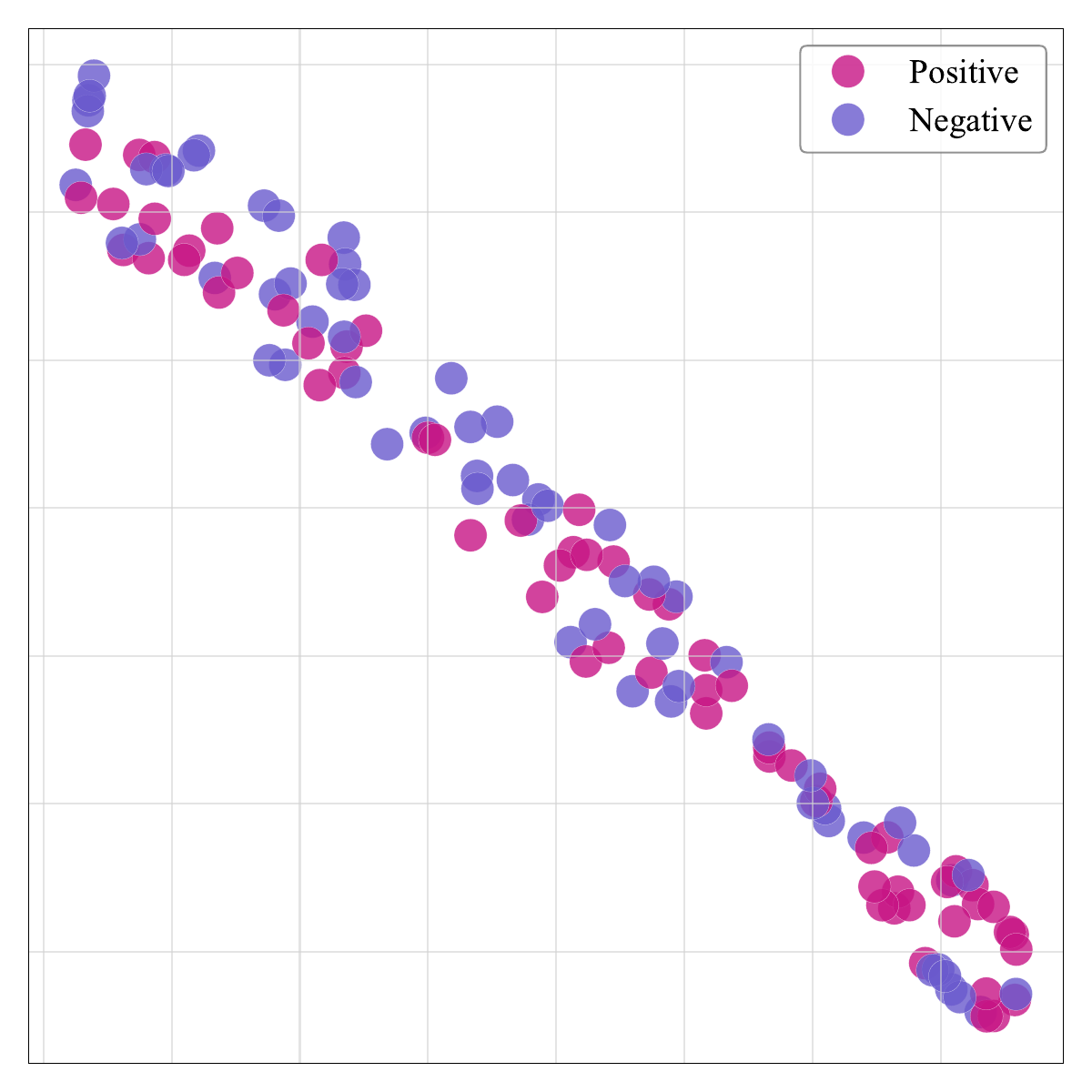}
			\end{subfigure}
			
			\caption{Visualization of feature disentanglement on the four datasets.}
			\label{fig:tsne_visualization_16grid_lda_sne}
		\end{figure*}

		We visualize the dual-stream feature distributions across four datasets in Figure~\ref{fig:tsne_visualization_16grid_lda_sne}. On smaller datasets (MOSI and CH-SIMS), the focus stream exhibit strong linear separability via distinct bimodal peaks, while the ambient stream remains chaotically entangled, successfully absorbing background states. Conversely, on larger-scale datasets (MOSEI and CH-SIMS v2), the focus stream shows significant distributional overlap between sentiment polarities. This overlap, particularly pronounced in CH-SIMS v2, directly reveals the performance bottleneck preventing state-of-the-art results on this specific benchmark. And these finding indicate that scaling our disentanglement mechanism to handle highly diverse, large-scale dataset remains a vital direction for future work.
	
	\subsection{Analysis of Fine-grained Sentiment Prediction}\label{app:confusionmatrix_ablation}
	
\begin{figure*}[b]
			\centering
			\setlength{\tabcolsep}{1pt} 
			
			\makebox[0.24\linewidth]{\small \textbf{MOSI}} \hfill
			\makebox[0.24\linewidth]{\small \textbf{MOSEI}} \hfill
			\makebox[0.24\linewidth]{\small \textbf{CH-SIMS}} \hfill
			\makebox[0.24\linewidth]{\small \textbf{CH-SIMS v2}}
			
			\vspace{2pt} 
			
			\begin{subfigure}[b]{0.24\linewidth}
				\includegraphics[width=\textwidth]{./fig/confusion_matrix_class7_qwen_MOSI.pdf}
			\end{subfigure}\hfill
			\begin{subfigure}[b]{0.24\linewidth}
				\includegraphics[width=\textwidth]{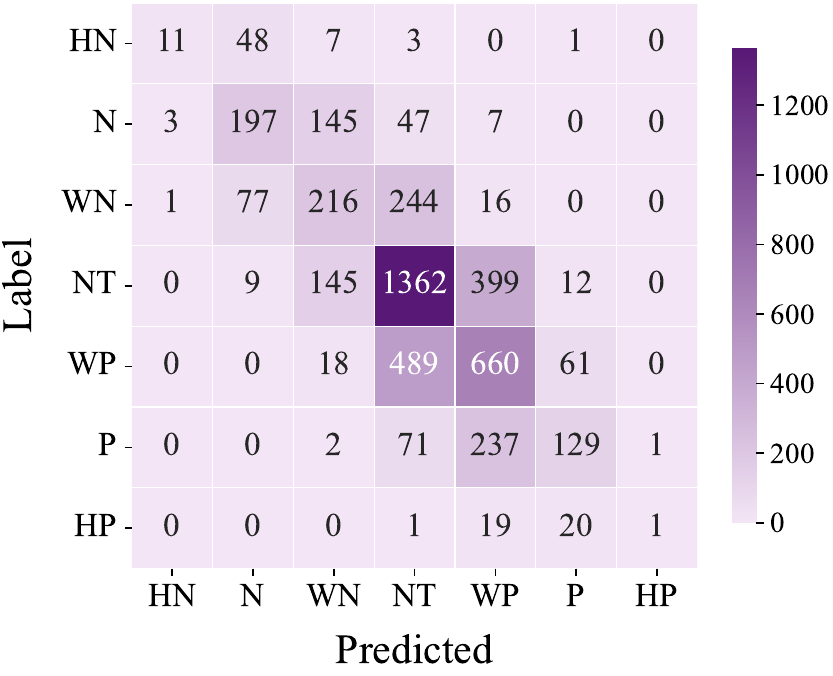}
			\end{subfigure}\hfill
			\begin{subfigure}[b]{0.24\linewidth}
				\includegraphics[width=\textwidth]{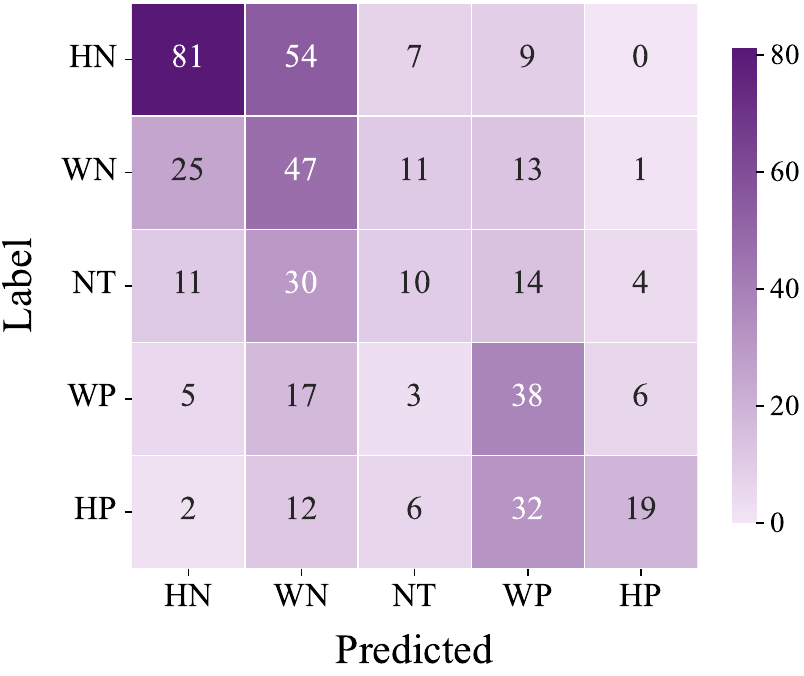}
			\end{subfigure}\hfill
			\begin{subfigure}[b]{0.24\linewidth}
				\includegraphics[width=\textwidth]{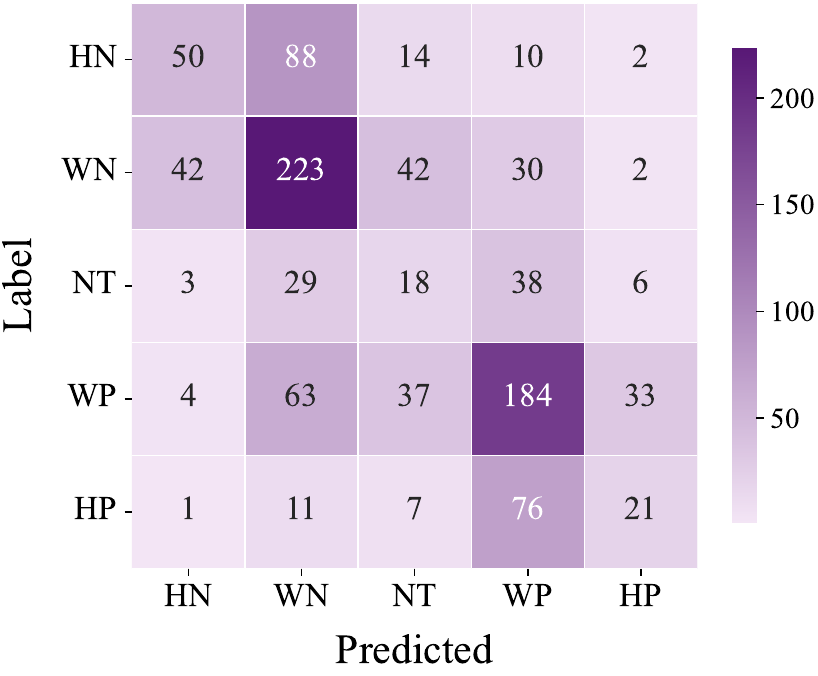}
			\end{subfigure}
			
			\vspace{2pt} 
			
			\begin{subfigure}[b]{0.24\linewidth}
				\includegraphics[width=\textwidth]{./fig/accuracy_per_class7_qwen_MOSI.pdf}
			\end{subfigure}\hfill
			\begin{subfigure}[b]{0.24\linewidth}
				\includegraphics[width=\textwidth]{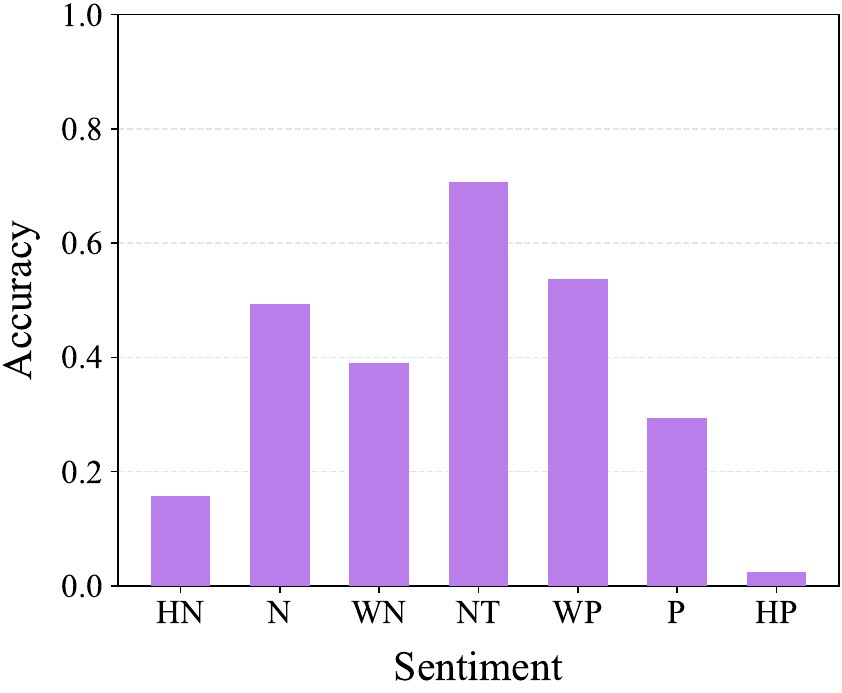}
			\end{subfigure}\hfill
			\begin{subfigure}[b]{0.24\linewidth}
				\includegraphics[width=\textwidth]{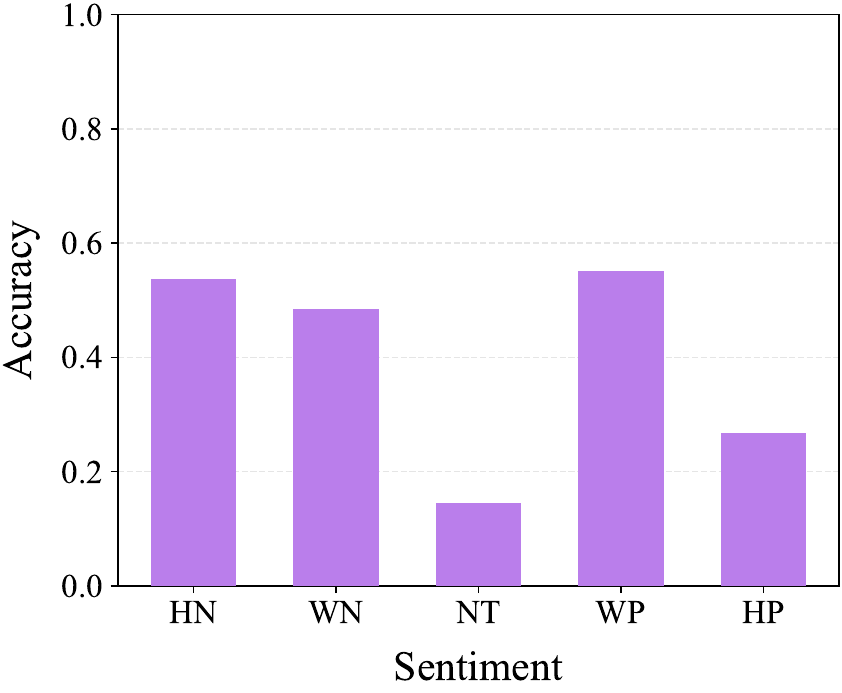}
			\end{subfigure}\hfill
			\begin{subfigure}[b]{0.24\linewidth}
				\includegraphics[width=\textwidth]{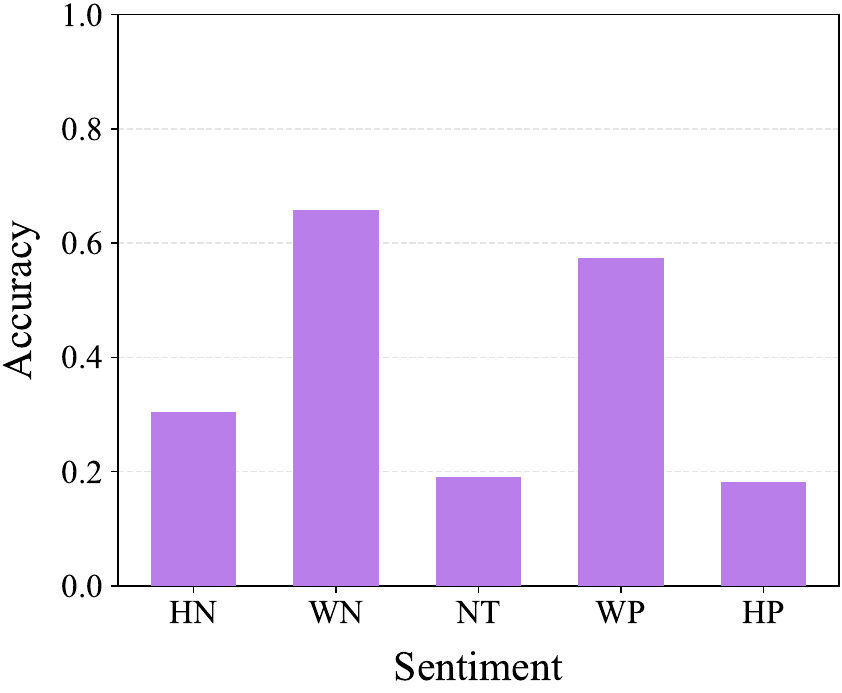}
			\end{subfigure}
			
			\vspace{8pt}
			
			\caption{Confusion matrix and per-class accuracy for 7-class sentiment prediction on the four datasets. HN: Highly Negative; N: Negative; WN: Weakly Negative; NT: Neutral; WP: Weak Positive; P: Positive; HP: Highly Positive.}
			\label{fig:tsne_visualization_8grid_cf_pclass}
		\end{figure*}

		To provide a deeper understanding of the model’s predictive behavior, we visualize the confusion matrix and per-class accuracies across all four datasets in Figure~\ref{fig:tsne_visualization_8grid_cf_pclass}. Consistent with our observations on MOSI, the MOSEI dataset exhibits a serve long-tail distribution where the model struggles significantly with the extreme polarities ("HN" and "HP") due to data sparsity. Interestingly, on the CH-SIMS and CH-SIMS v2 datasets, the bottleneck shifts to "NT" class, which yields the lowest accuracy. This discrepancy is likely because neutral expression in these datasets are highly nuanced and easily misclassified into neighboring weak sentiment classes (e.g., WN or WP). These results indicate that the inherent class imbalance constrains the theoretical upper bound of sentiment prediction, highlighting long-tail multimodal learning as a critical direction for our future work.

		\begin{table*}[b] 
			\centering
			\caption{Complete prompt list of semantic sentences used for query initialization on MOSI and MOSEI datasets.}
			\label{tab:tasq_prompts_mosi}
			
			\begin{PromptBox}{Complete Prompt List of Semantic Sentences for MOSI / MOSEI}
				
				\textbf{\# Video Modality}
				
				\begin{itemize}[leftmargin=1em, label=\textcolor{blue!60!black}{\texttt{>}}, itemsep=2pt]
					
					\item \texttt{a face showing clear positive emotion (smile, joyful)} 
					\hfill \textcolor{gray!80}{\texttt{\# Positive Valence}}
					
					\item \texttt{a face showing clear negative emotion (frown, sadness, anger)} 
					\hfill \textcolor{gray!80}{\texttt{\# Negative Valence}}
					
					\item \texttt{a facial expression with moderate intensity (subtle emotion)} 
					\hfill \textcolor{gray!80}{\texttt{\# Low Intensity}}
					
					\item \texttt{a sudden change in facial expression or head movement} 
					\hfill \textcolor{gray!80}{\texttt{\# Temporal Dynamics}}
					
					\item \texttt{a speaker gesturing emphatically (strong body language)} 
					\hfill \textcolor{gray!80}{\texttt{\# Body Language}}
					
					\item \texttt{a speaker with restrained micro-expressions or tension} 
					\hfill \textcolor{gray!80}{\texttt{\# Micro-expression}}
					
					\item \texttt{a social interaction cue (eye contact, mutual gaze)} 
					\hfill \textcolor{gray!80}{\texttt{\# Social Cue}}
					
					\item \texttt{a neutral face or resting expression (no strong emotion)} 
					\hfill \textcolor{gray!80}{\texttt{\# Neutral}}
					
				\end{itemize}
				
				\vspace{0.3cm} 
				\textbf{\# Audio Modality}
				
				\begin{itemize}[leftmargin=1em, label=\textcolor{blue!60!black}{\texttt{>}}, itemsep=2pt]
					
					\item \texttt{a voice with a warm, friendly tone (positive prosody)} 
					\hfill \textcolor{gray!80}{\texttt{\# Positive Prosody}}
					
					\item \texttt{a voice with a tense or harsh tone (angry or stressed)} 
					\hfill \textcolor{gray!80}{\texttt{\# Negative Prosody}}
					
					\item \texttt{low-intensity prosodic variation (subtle change in pitch)} 
					\hfill \textcolor{gray!80}{\texttt{\# Subtle Variation}}
					
					\item \texttt{a sudden change in pitch or loudness (surge or exclamation)} 
					\hfill \textcolor{gray!80}{\texttt{\# Transient Event}}
					
					\item \texttt{vocal emphasis or strong stress on words (emphatic speech)} 
					\hfill \textcolor{gray!80}{\texttt{\# Emphatic Speech}}
					
					\item \texttt{vocal micro-intonations like sighs, breathiness, or pauses} 
					\hfill \textcolor{gray!80}{\texttt{\# Paralinguistics}}
					
					\item \texttt{prosodic pattern of uncertainty or hesitation (uh, um)} 
					\hfill \textcolor{gray!80}{\texttt{\# Hesitation}}
					
					\item \texttt{a neutral, flat prosodic contour (monotone)} 
					\hfill \textcolor{gray!80}{\texttt{\# Neutral}}
					
				\end{itemize}				
				
			\end{PromptBox}
			
		\end{table*}
		
		\begin{CJK}{UTF8}{gbsn}
			
			\begin{table*}[b] 
				\centering
				\caption{Complete prompt list of semantic sentences used for query initialization on CH-SIMS and CH-SIMS v2 datasets.}
				\label{tab:tasq_prompts_sims}
				
				\begin{PromptBox}{Complete Prompt List of Semantic Sentences for CH-SIMS / CH-SIMS v2}
					
					\textbf{\# Video Modality}
					
					\begin{itemize}[leftmargin=1em, label=\textcolor{blue!60!black}{\texttt{>}}, itemsep=2pt]
						
						\item \textnormal{嘴角上扬、嘴唇张开或明显的嘴部肌肉运动} 
						\hfill \textcolor{gray!80}{\textnormal{\#笑容、说话、嘲讽}}
						
						\item \textnormal{眉毛皱起、紧锁或前额肌肉的剧烈收缩} 
						\hfill \textcolor{gray!80}{\textnormal{\# 愤怒、思考、痛苦}}
						
						\item \textnormal{眼睛睁大、眉毛上挑或眼周肌肉的扩张} 
						\hfill \textcolor{gray!80}{\textnormal{\# 惊讶、恐惧、强调}}
						
						\item \textnormal{面部肌肉松弛、平坦或缺乏明显的纹理变化} 
						\hfill \textcolor{gray!80}{\textnormal{\# 中性、发呆、冷漠}}
						
						\item \textnormal{头部的大幅度转动、点头或剧烈的晃动} 
						\hfill \textcolor{gray!80}{\textnormal{\# 否认、肯定、激动}}
						
						\item \textnormal{单侧嘴角抽动、眼神斜视或不对称的面部表情} 
						\hfill \textcolor{gray!80}{\textnormal{\# 轻蔑、不屑}}
						
						\item \textnormal{明显的手部挥舞、指点或身体姿态的前后移动} 
						\hfill \textcolor{gray!80}{\textnormal{\# 肢体语言、情绪强度}}
						
						\item \textnormal{极短时间内发生的快速面部抽动或微表情} 
						\hfill \textcolor{gray!80}{\textnormal{\# 压抑、掩饰、紧张}}
						
					\end{itemize}
					
					\vspace{0.3cm} 
					\textbf{\# Audio Modality}
					
					\begin{itemize}[leftmargin=1em, label=\textcolor{blue!60!black}{\texttt{>}}, itemsep=2pt]
						
						\item \textnormal{音高较高、频率尖锐或上升的语调模式} 
						\hfill \textcolor{gray!80}{\textnormal{\# 兴奋、尖叫、质问}}
						
						\item \textnormal{音量巨大、能量爆发或带有冲击力的声音} 
						\hfill \textcolor{gray!80}{\textnormal{\# 愤怒、强调、惊叹}}
						
						\item \textnormal{低沉、沙哑、微弱或频率较低的声音特质} 
						\hfill \textcolor{gray!80}{\textnormal{\# 悲伤、疲惫、阴沉}}
						
						\item \textnormal{语速极快、连贯紧凑且没有停顿的说话方式} 
						\hfill \textcolor{gray!80}{\textnormal{\# 紧张、急切、激辩}}
						
						\item \textnormal{明显的长时间停顿、沉默或极慢的语速} 
						\hfill \textcolor{gray!80}{\textnormal{\# 犹豫、尴尬、思考}}
						
						\item \textnormal{呼吸声、叹气声、笑声或非语言的喉音} 
						\hfill \textcolor{gray!80}{\textnormal{\# 无奈、轻蔑、潜台词}}
						
						\item \textnormal{节奏平稳、单调且缺乏起伏的机械式发音} 
						\hfill \textcolor{gray!80}{\textnormal{\# 中性、陈述、冷漠}}
						
						\item \textnormal{音调突然转折、断裂或不稳定的颤音} 
						\hfill \textcolor{gray!80}{\textnormal{\# 哭腔、破音、极度激动}}
						
					\end{itemize}				
					
				\end{PromptBox}
				
			\end{table*}
			
		\end{CJK}
		
\section{Limitations}\label{app:limitation}
	
	Despite the demonstrated effectiveness, there exist some limitations: 
	
	\begin{enumerate}
		
		\item \textbf{Interpretability}: We currently rely on LLM reasoning for sentiment classification. However, classification labels alone do not fully reveal the internal decision logic. Future work could explore generating textual explanations to enhance interpretability and transparency.
		
		\item \textbf{Missing-Modality Scenarios}: Our method assumes complete modality availability. In real-world applications, data corruption often leads to missing signals. Improving robustness under missing-modality scenarios remains a critical direction for future work.
		
	\end{enumerate}
	
\end{document}